%% file: main.tex
\documentclass[sigconf,screen,nonacm]{acmart}

\usepackage{booktabs}
\usepackage{siunitx}
\usepackage{threeparttable}
\usepackage{multirow}
\usepackage{makecell}
\usepackage{amsthm}
\usepackage{subcaption}
\usepackage{bbm}
\usepackage{enumitem}

\renewcommand\footnotetextcopyrightpermission[1]{}
\newcommand\blfootnote[1]{%
  \begingroup
  \renewcommand\thefootnote{}%
  \footnote{#1}%
  \addtocounter{footnote}{-1}%
  \endgroup
}

\newtheorem{assumption}{Assumption}[section]
\newtheorem{proposition}{Proposition}[section]

\AtBeginDocument{%
  }

\begin{document}

\title{SafeImpute: Reliable Clinical Data Imputation \\ via Conformal Selection}

\author{Xinrui He}
\affiliation{%
  \institution{University of Illinois Urbana-Champaign}
  \city{Champaign}
  \state{IL}
  \country{USA}
}
\email{xhe33@illinois.edu}

\author{Mengting Ai}
\affiliation{%
  \institution{University of Illinois Urbana-Champaign}
  \city{Champaign}
  \state{IL}
  \country{USA}
}
\email{mai10@illinois.edu}

\author{Junting Wang}
\affiliation{%
  \institution{University of Illinois Urbana-Champaign}
  \city{Champaign}
  \state{IL}
  \country{USA}
}
\email{junting3@illinois.edu}

\author{Curtiss B. Cook}
\affiliation{%
  \institution{Mayo Clinic}
  \city{Scottsdale}
  \state{AZ}
  \country{USA}
}
\email{cook.curtiss@mayo.edu}

\author{Jingrui He}
\affiliation{%
  \institution{University of Illinois Urbana-Champaign}
  \city{Champaign}
  \state{IL}
  \country{USA}
}
\email{jingrui@illinois.edu}

\renewcommand{\shortauthors}{Xinrui He, Mengting Ai, Junting Wang, Curtiss B. Cook and Jingrui He}
\newcommand{\model}{{{\tt SafeImpute}}}

\begin{abstract}
Clinical care often relies on key laboratory indicators, yet real-world patient visits are sparse and tests are ordered irregularly, leading to pervasive missingness. While many imputation methods improve average accuracy, they provide limited guidance on which imputed values are reliable enough for high-stakes downstream use. In this work, we study reliable clinical imputation, aiming to produce accurate imputations while selectively releasing the reliable results, with statistical control over clinically unacceptable errors. To achieve this goal, we propose \model{}, a reliable imputation framework for irregular and sparse clinical longitudinal records. \model{} constructs an event graph that captures both intra-patient temporal trajectories and inter-patient clinical similarity, and learns imputations with a two-relation GNN and adaptive fusion, regularized by an auxiliary masked reconstruction objective. For reliability guarantees, \model{} converts a proxy risk score into conformal p-values and applies the Benjamini--Hochberg procedure to control the false discovery rate (FDR) of unacceptable errors among released imputations at a user-specified tolerance. Experiments on our Mayo Clinic data, the public MIMIC-III and MIMIC-IV datasets show that \model{} achieves strong imputation accuracy while providing reliable error control, outperforming diverse baselines in both standard imputation evaluation and FDR-controlled selective-release evaluation. Code is available at \url{https://github.com/Xinrui17/SafeImpute}.
\end{abstract}




\maketitle

\blfootnote{Accepted at the 32nd ACM SIGKDD Conference on Knowledge Discovery and Data Mining (KDD 2026). This is the authors' accepted manuscript. The final version of record will appear in the ACM Digital Library with DOI: 10.1145/3770855.3817967.}

\input{sec/1_intro}

\input{sec/2_preliminary}
\input{sec/3_method}

\input{sec/4_experiments}

\input{sec/5_related_work}
\input{sec/6_conclusion}

\bibliographystyle{ACM-Reference-Format}
\bibliography{main}
\appendix

\input{sec/appendix_1}
\input{sec/appendix_2}
\input{sec/appendix_5}

\input{sec/appendix_4}

\end{document}

%% file: sec/1_intro.tex
\section{Introduction}
In clinical care, key indicators, e.g., Hemoglobin A1c (HbA1c) for diabetes management \cite{Ban31122025}, are crucial for disease monitoring and treatment adjustment \cite{sherwani2016significance}, yet patient visits are irregular, and labs are often not ordered at every visit, resulting in pervasive missingness in longitudinal records. Recent years have seen growing interest in clinical data imputation, with methods ranging from statistical models \cite{kim2004reuse, troyanskaya2001missing,vafaei2025flexible} to deep generative models and representation learning \cite{grape, du2024remasker, zhang2025diffputer,beaulieu2017missing}. 
Despite this progress, most approaches emphasize overall accuracy and provide limited guarantees on error control in high-stakes use, where unreliable results can directly mislead diagnosis and treatment adjustment.

Clinical deployment, therefore, raises a question beyond overall accuracy: which imputed results are reliable enough to support clinical decision-making, and how can we provide explicit control over imputation errors? This motivates reliable clinical imputation, which augments imputation with uncertainty or reliability assessment, thereby producing imputations with quantitative error control. Related quality-control frameworks have been studied in high-stakes prediction, including Bayesian methods \cite{blei2017variational,wu2021quantifying}, imprecise probability \cite{augustin2014introduction}, and conformal prediction \cite{angelopoulos2024theoretical}. Several works have begun to apply these ideas in clinical settings~\cite{lu2022fair, giustinelli2022precise, kapuria2024novel,millar2024uncertainty}, such as conformalized models for genomic medicine \cite{genomic_prediction} and SAFER \cite{shen2025safer}, which uses conformal prediction to provide statistical false discovery rate (FDR) control for treatment prediction. However, risk-controlled clinical imputation remains underexplored, particularly for irregular and sparsely observed patient records \cite{bates2023testing,jin2023selection}.

Building on this gap, we focus on two challenges in reliable clinical data imputation. \textbf{(i) Modeling irregular, sparse, and heterogeneous longitudinal records.} Real-world patient visits usually occur at uneven intervals, and lab tests are ordered opportunistically, yielding highly sparse and heterogeneous measurement patterns. As a result, intra-patient temporal evidence is often limited, while informative signals may instead come from cross-patient analogies. This calls for an imputation model that can exploit both the limited temporal trajectory and relational structure across patients under severe missingness. \textbf{(ii) Reliable imputation with statistical error control.} In high-stakes settings, the goal is not only to minimize average error, but to determine which imputations are reliable enough to act on, and provide a quantitative guarantee that clinically unacceptable errors are rare among the released results. This requires uncertainty quantification for each imputed entry and a statistically grounded selection procedure that can control the error rate when releasing many imputations.

To address these challenges, we propose \model, a reliable imputation framework for irregular and sparse clinical records that couples a strong event graph imputer with alse discovery rate (FDR) control. Specifically, to address the irregular and sparse nature of clinical visits and measurements, we construct an event graph with temporal edges that capture intra-patient trajectories and trend-aware value edges that connect clinically similar events to capture inter-patient analogies. On this event graph, we learn event representations with a two-relation GNN that performs relation-specific message passing and fuses temporal and value signals via adaptive gating. To ensure reliability in high-stakes settings, \model{} computes a proxy risk score for each imputed entry and calibrates it into conformal p-values, then applies the Benjamini--Hochberg procedure to select a subset of imputations while controlling the FDR of clinically unacceptable errors at a user-specified tolerance level. Together, \model{} produces accurate imputations and selectively releases them with statistical error control.

The major contributions of this paper are summarized as follows:
\begin{itemize}
    \item We formulate the problem of reliable clinical data imputation, where the goal is not only to achieve accurate imputation but also to selectively release reliable results with statistical control over clinically unacceptable errors.
     \item We propose \model{}, a reliable imputation framework that (i) represents irregular clinical visits as an event graph, where temporal edges capture intra-patient trajectories and similarity edges connect clinically related visits across patients, and learns imputations using a two-relation GNN with adaptive fusion; and (ii) provides reliability control by calibrating a proxy risk score into conformal p-values and applying the Benjamini--Hochberg procedure to control the false discovery rate among released imputations at a user-specified tolerance.
    \item Extensive experiments on Mayo Clinic dataset and the MIMIC-III/IV datasets demonstrate that \model{} achieves strong imputation accuracy while providing reliable error control.
\end{itemize}



%% file: sec/2_preliminary.tex
\section{Preliminary}

\subsection{Clinical Imputation}
\label{sec:prelim_ehr_imputation}

In many clinical decision-making scenarios, decisions are driven by a small set of decision-critical laboratory markers that are directly relevant for diagnosis and treatment planning, rather than by completing every missing measurement.
Motivated by this, we focus on imputing a designated decision-critical target, such as HbA1c for diabetes monitoring and treatment adjustment. We consider irregular, multi-lab longitudinal clinical records, where each patient has an irregular set of clinical visits over time and only a subset of laboratory tests may be measured at each visit.
We treat each visit as an event node that aggregates the laboratory tests recorded during that visit.
Let $\mathcal{V}$ denote the set of all event nodes across patients.
For node $i\in\mathcal{V}$, let $p_i$ denote its patient identity and $t_i$ its timestamp.
Let $\mathcal{L}$ be the set of laboratory variables together with the available attributes of the patient.
We denote the value of variable $\ell\in\mathcal{L}$ at node $i$ by $z_{i,\ell}$. Formally, given partially observed multi-lab inputs $\{z_{i,\ell}\}_{\ell\in\mathcal{L}}$, our goal is to impute only the designated target $y_{i}$ at event nodes where it is missing.

\subsection{Conformal Selection}

Conformal selection (CS) \cite{jin2023selection} is a model-agnostic framework for selecting a subset of test instances while controlling the false discovery rate (FDR) \cite{benjamini1995controlling} in finite samples.

CS formulates one hypothesis test for each candidate test sample.
Let $\{(x_i,y_i)\}_{i=1}^n$ denote the calibration samples and $\{x_{n+j}\}_{j=1}^m$ denote the test candidates.
For each test candidate $j\in[m]$, CS considers the hypotheses
\begin{equation}
H_{0j}:~ y_{n+j}\le c
\label{eq:cs_hypothesis}
\end{equation}
Rejecting $H_{0j}$ indicates that the $j$-th test candidate is deemed to exceed the threshold $c$.

\paragraph{Nonconformity scores and conformal p-values.}
CS relies on a nonconformity measure $J(\cdot,\cdot)$ that quantifies how atypical an observation is relative to the calibration data.
For calibration samples, the nonconformity scores are
\begin{equation}
J_i = J(x_i,y_i),\quad i=1,\dots,n.
\end{equation}
For a test candidate $x_{n+j}$, since $y_{n+j}$ is unobserved, CS replaces it by the threshold $c$ and computes
\begin{equation}
\widehat{J}_{n+j} = J(x_{n+j},c).
\end{equation}
The conformal p-value \cite{jin2023selection} for testing~\eqref{eq:cs_hypothesis} is then computed via a rank-based comparison between $\widehat{J}_{n+j}$ and $\{J_i\}_{i=1}^n$.
Intuitively, a smaller conformal p-value provides stronger evidence against $H_{0j}$.

\paragraph{BH procedure and FDR control.}
To determine the final selected subset, CS applies the Benjamini--Hochberg (BH) \cite{benjamini1995controlling} procedure to the set of conformal p-values $\{p_{n+j}\}_{j=1}^m$.
Let $p_{(1)}\le \cdots \le p_{(m)}$ be the sorted p-values and define
\begin{equation}
k^\star = \max\left\{k\in[m]:~ p_{(k)} \le \frac{k}{m}\,\alpha \right\},
\end{equation}
with the convention that \(k^\star=0\) if the set is empty. $\alpha\in(0,1)$ is the target FDR level.
BH rejects all hypotheses with $p_{n+j}\le p_{(k^\star)}$, yielding the selected set.
Under the standard validity conditions for conformal p-values (e.g., exchangeability), this procedure guarantees finite-sample FDR control at level $\alpha$.

%% file: sec/3_method.tex
\section{Method}

In this section, we present \model{}, including event-graph construction for irregular clinical records, two-relation GNN learning over temporal and trend-aware value edges, and the selection module that provides FDR-controlled selective imputation release.

\begin{figure*}[t!]
    \centering
    \includegraphics[width=0.9\linewidth]{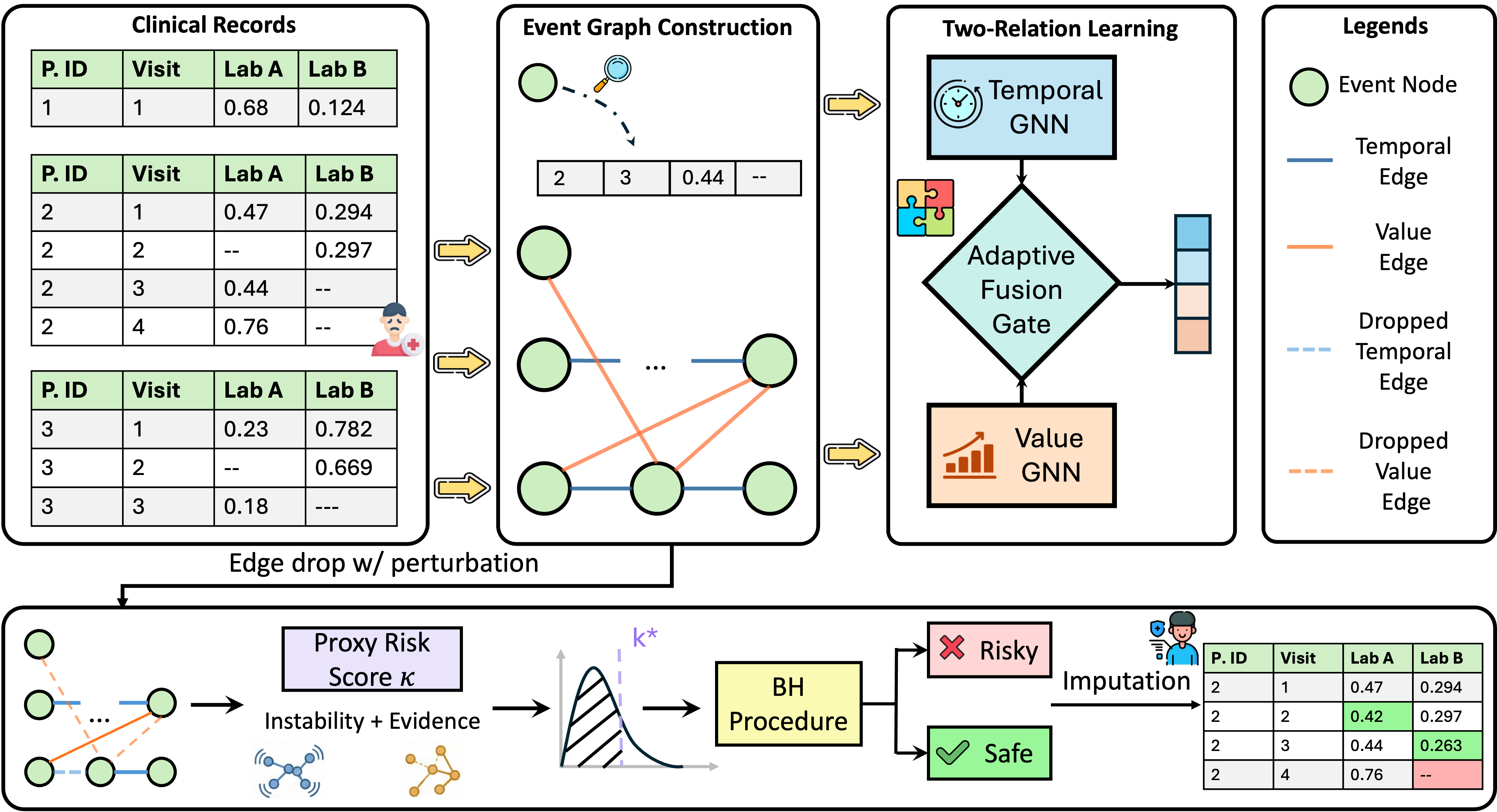}
    \caption{\model{} overview for reliable clinical imputation. We convert irregular and sparse longitudinal records into an event graph with temporal edges and trend-aware value edges, then learn with a two-relation GNN combined with an adaptive fusion gating network to impute missing labs. To achieve FDR control over clinically unacceptable errors, we compute a proxy risk score that combines perturbation-induced instability with evidence degree, convert it to conformal p-values, and apply the Benjamini–Hochberg procedure to selectively release safe imputations.}
    \label{fig:main_fig}
\end{figure*}

\subsection{Event Graph Construction}

\paragraph{Event Node.} 

We treat each patient visit as an event node. To preserve as much event-level information as possible under irregular missingness, we encode both available lab values and missingness patterns in the node feature. Specifically, for each lab $\ell\in\mathcal{L}$ at event node $i$, we define an observation indicator
\begin{equation}
m_{i,\ell}=\mathbb{I}\big[z_{i,\ell}\ \text{is observed}\big].
\end{equation}
We then represent each event by concatenating lab values with their indicators:
\begin{equation}
\mathbf{x}_i=\Big[\, z_{i,1}, m_{i,1},\ z_{i,2}, m_{i,2},\ \ldots,\ z_{i,|\mathcal{L}|}, m_{i,|\mathcal{L}|},\ g_i \Big],
\end{equation}
where unobserved $z_{i,\ell}$ are set to $0$ only for feature construction. To mitigate scale differences across lab values, we normalize the lab values.

With event nodes, we construct an event graph $\mathcal{G}=(\mathcal{V},\mathcal{E}_t\cup\mathcal{E}_v)$ with two types of edges
 that capture two complementary signals:
\emph{intra-patient trajectories} and \emph{inter-patient analogies}.
Specifically, we include
(1) intra-patient temporal edges $\mathcal{E}_t$ that connect consecutive events of the same patient, preserving local temporal continuity; and
(2) inter-patient trend-aware value edges $\mathcal{E}_v$ that connect events among patients using lab value-level and trend-level information, enabling information sharing between clinically similar event contexts. We describe the construction of these two edge types below.

\paragraph{Temporal Edges.}
To preserve intra-patient trajectories, for each patient $p$ we sort its event nodes by time.
Let $\text{next}(i)$ denote the immediate subsequent event of the same patient after node $i$.
We add a temporal edge between consecutive events if the time gap is within a threshold:
\begin{equation}
\mathcal{E}_t = \left\{(i,j)\ \middle|\ p_i=p_j,\ j=\text{next}(i),\ 0 < t_j-t_i \le \Delta_t \right\}.
\end{equation}

\paragraph{Trend-aware Value Edges.}
Temporal edges capture within-patient continuity but do not facilitate information sharing across patients.
To model inter-patient analogies, we additionally connect events that exhibit similar lab levels and temporal trends.

For each node $i$ and lab $\ell$, let $r_{i,\ell}$ denote the most recent prior event of patient $p_i$ before $t_i$ where lab $\ell$ is observed.
If such an event does not exist, we set $r_{i,\ell}=\varnothing$.
We define the per-lab trend as the rate of change from the last available observation:
\begin{equation}
\delta_{i,\ell}=
\begin{cases}
\dfrac{z_{i,\ell}-z_{r_{i,\ell},\ell}}{(t_i-t_{r_{i,\ell}})+\epsilon}, & r(i,\ell)\neq\varnothing,\ t_i>t_{r_{i,\ell}},\\[6pt]
\text{missing}, & \text{otherwise}.
\end{cases}
\end{equation}

For a pair of nodes $(i,j)$, the sets of commonly observed labs for values and trends are defined as:
\begin{equation}
\mathcal{L}_{ij} = \left\{\ell\in\mathcal{L}\mid z_{i,\ell}\ \text{and}\ z_{j,\ell}\ \text{are observed}\right\},
\end{equation}

\begin{equation}
\mathcal{L}^{\delta}_{ij} = \left\{\ell\in\mathcal{L}\mid \delta_{i,\ell}\ \text{and}\ \delta_{j,\ell}\ \text{are observed}\right\}.
\end{equation}
To avoid node similarity being biased by the number of commonly observed labs, we compute RMS-normalized Euclidean distances over the common dimensions as value and trend similarities:
\begin{align}
d_v(i,j) &= \sqrt{\frac{1}{|\mathcal{L}_{ij}|}\sum_{\ell\in\mathcal{L}_{ij}}\left(\hat{z}_{i,\ell}-\hat{z}_{j,\ell}\right)^2},\\
d_\delta(i,j) &= \sqrt{\frac{1}{|\mathcal{L}^{\delta}_{ij}|}\sum_{\ell\in\mathcal{L}^{\delta}_{ij}}\left(\hat{\delta}_{i,\ell}-\hat{\delta}_{j,\ell}\right)^2},
\end{align}

where $\hat{z}_{i,\ell}$ and $\hat{\delta}_{i,\ell}$ denote the normalized value and trend features, respectively. We connect $i$ and $j$ if both the value and trend distance criteria are satisfied:
\begin{equation}
\mathcal{E}_v =\{(i,j)| d_v(i,j) \le \tau_v, d_\delta(i,j)\le \tau_\delta\}.
\end{equation}
To avoid isolated nodes under sparse observations, for any node $i$ with $\mathcal{L}^{\delta}_{ij}=\emptyset$ for all $j$, we connect $i$ to its nearest value-based neighbor.

\subsection{Two-Relation Learning}

To learn event representations that jointly capture intra-patient progression and inter-patient similarity, we propose a two-relation graph neural network (GNN) that performs separate message passing over temporal edges $\mathcal{E}_t $ and value edges $\mathcal{E}_v$, followed by adaptive fusion at each layer.

Let $\mathbf{h}_i^{(0)}=\mathbf{x}_i$. At layer $k$, we compute relation-wise messages for each node $i$:
\begin{align}
\mathbf{m}^{(k)}_{t,i} &= \mathrm{GNN}_t^{(k)}(\mathbf{H}^{(k-1)};\mathcal{E}_t)_i, \\
\mathbf{m}^{(k)}_{v,i} &= \mathrm{GNN}_v^{(k)}(\mathbf{H}^{(k-1)};\mathcal{E}_v)_i,
\end{align}
where $\mathrm{GNN}_t^{(k)}$ and $\mathrm{GNN}_v^{(k)}$ are GCN layers with independent parameters:
\begin{equation}
\mathrm{GNN}(\mathbf{H};\mathcal{E}) = \sigma\!\left(\tilde{\mathbf{D}}^{-\frac12}\tilde{\mathbf{A}}\tilde{\mathbf{D}}^{-\frac12}\mathbf{H}\mathbf{W}\right),
\end{equation}
where $\tilde{\mathbf{A}}$ is the adjacency with self-loops and $\tilde{\mathbf{D}}$ is the corresponding degree matrix, $\mathbf{W}$ is the learnable weight matrix and $\sigma(\cdot)$ is the activation function.

We fuse the two relation-specific messages using a learnable gate computed for each node at each layer:
\begin{equation}
\alpha_i^{(k)} =
\mathrm{sigmoid}\!\left(
\mathbf{w}^{(k)\top}
\big(\mathbf{m}^{(k)}_{t,i}+\mathbf{m}^{(k)}_{v,i}\big)
\right),
\end{equation}
\begin{equation}
\mathbf{h}_i^{(k)} =
\alpha_i^{(k)}\mathbf{m}^{(k)}_{t,i}
+
\big(1-\alpha_i^{(k)}\big)\mathbf{m}^{(k)}_{v,i}.
\end{equation}

After the final layer $K$, we apply a multi-output linear prediction head:
\begin{equation}
\hat{\mathbf z}_i = \mathbf{W}_o \mathbf{h}_i^{(K)} + \mathbf{b}_o,
\end{equation}
where each output dimension corresponds to one laboratory variable. The prediction for the designated target measurement is denoted by $\tilde{y}_i$, i.e., the target dimension of $\hat{\mathbf z}_i$.

Let $\Omega_{\mathrm{train}}$ be the set of training nodes with observed target labels $y_i$. We train the main target prediction by minimizing
\begin{equation}
\mathcal{L}_{\mathrm{target}} =
\frac{1}{|\Omega_{\mathrm{train}}|}
\sum_{i\in\Omega_{\mathrm{train}}}
(\tilde{y}_i-y_i)^2.
\end{equation}

Although the final task is to impute the designated target measurement, we further regularize the event representations by reconstructing randomly masked non-target laboratory values. At each epoch, we randomly mask a small fraction of observed non-target labs from the input and ask the model to reconstruct them from the learned event representation. Let $\mathcal{M}_\ell$ be the set of masked node--lab pairs for lab $\ell$. To reduce sensitivity to outlying lab values, we use the Huber reconstruction loss:
\begin{equation}
\mathcal{L}_{\mathrm{aux}}
=
\frac{1}{\sum_{\ell \ne y} |\mathcal{M}_\ell|}
\sum_{\ell \ne y}
\sum_{(i,\ell)\in \mathcal{M}_\ell}
\rho_{\eta}(\hat z_{i,\ell}-z_{i,\ell}),
\end{equation}
where
\begin{equation}
\rho_{\eta}(e)=
\begin{cases}
\frac{1}{2}e^2, & |e|\le \eta,\\
\eta\left(|e|-\frac{1}{2}\eta\right), & |e|>\eta.
\end{cases}
\end{equation}
The final training objective is
\begin{equation}
\mathcal{L}
=
\mathcal{L}_{\mathrm{target}}
+
\lambda \mathcal{L}_{\mathrm{aux}}.
\end{equation}

\subsection{Conformal Selection and FDR Control}
In clinical settings, imputed laboratory values may directly influence diagnosis, prognosis assessment, and treatment planning.
We therefore apply conformal selection with false discovery rate (FDR) control to retain only reliable imputations: among the released imputations, the expected fraction of clinically unacceptable errors is controlled at a user-specified level.

\paragraph{Risk hypothesis.}
Let $\hat{y}_i$ be the model prediction at node $i$ and $y_i$ be the ground-truth label (available only on calibration).
We define the true error on labeled points as
\begin{equation}
\kappa(i) = |y_i-\hat{y}_i|.
\end{equation}
Given a user-specified tolerance $\delta>0$ for the target variable, we call node $i$ \emph{risky} if $\kappa(i)\ge \delta$.
For each test-time node $j$ (with unknown $y_j$), we consider the null hypothesis
\begin{equation}
H_j:\ \kappa(j)\ge \delta
\qquad \text{(the imputation at $j$ is risky)}.
\label{eq:hypothesis-risky}
\end{equation}

\paragraph{Proxy risk score.}
Since $\kappa(j)$ is unobservable at test time, conformal selection requires a label-free proxy score that tends to be larger for risky imputations.
We therefore define a proxy risk score by combining two complementary signals:
prediction instability under structure-preserving perturbations, capturing sensitivity to small changes in relational evidence; and an evidence penalty, capturing the scarcity of relational support even when predictions appear stable.

To compute prediction instability, we perform margin-weighted edge perturbations that reflect how strongly each edge is supported by the edge construction rules.
For each edge $e$, we compute a normalized margin $m(e)\in[0,1]$ to the threshold and convert it into an edge-specific keep probability
\begin{equation}
\pi(e) \;=\; \pi_{\min} + (\pi_{\max}-\pi_{\min})\cdot m(e)^{\gamma},
\label{eq:keep_prob}
\end{equation}
where $\pi_{\min},\pi_{\max}\in(0,1)$ and $\gamma>0$.
Edges that barely satisfy the threshold have small $m(e)$ and are dropped more often, while strongly supported edges have large $m(e)$ and are kept with higher probability.

For a temporal edge $e=(i,j)\in\mathcal{E}_t$:
\begin{equation}
m_t(e)\;=\;1-\min\!\left\{ \frac{\Delta t_{ij}}{\Delta_t},\,1\right\}.
\label{eq:temporal_margin}
\end{equation}
and for a value edge $e=(i,j)\in\mathcal{E}_v$, 

\begin{equation}
s_v(e)\;=\;\max\!\left\{\frac{d^{v}_{ij}}{\tau_{v}},\,\frac{d^{\delta}_{ij}}{\tau_{\delta}}\right\},
\quad
m_v(e)=1-\min\{s_v(e),1\}
\label{eq:value_margin}
\end{equation}
If the trend is unavailable for at least one endpoint, we use value-only strength $s_v(e)=d^v_{ij}/\tau_v$.

Let $\mathcal{G}^{(1)},\ldots,\mathcal{G}^{(K)}$ be $K$ independent perturbations obtained by sampling each edge $e$ according to $\mathrm{Bernoulli}(\pi(e))$.
Let $\hat{y}_i^{(k)}$ be the model prediction for node $i$ when running inference on $\mathcal{G}^{(k)}$.
We define the prediction-instability score as the dispersion of these predictions:
\begin{equation}
S_{\mathrm{pred}}(i)
=\mathrm{Std}\Big(\{\hat{y}_i^{(k)}\}_{k=1}^{K}\Big),
\label{eq:spred}
\end{equation}
where $\mathrm{Std}(\cdot)$ denotes the standard deviation across perturbations.

Beyond perturbation sensitivity, reliability also depends on the amount of relational evidence supporting a node.
When the constructed relations are sparse for a node (hence small degree), there is limited information propagation from other events, and the imputation is driven by a weak neighborhood signal.
To reflect this evidence scarcity, we incorporate a degree-based penalty that downweights nodes with insufficient temporal/value support, independent of the instability term.

Concretely, let $\deg_t(i)$ and $\deg_v(i)$ denote the degrees of node $i$ in the temporal and value graphs, respectively.
We define the evidence penalty as
\begin{equation}
S_{\mathrm{evid}}(i)
=
\frac{1}{\sqrt{\deg_t(i)+1}}
+
\frac{1}{\sqrt{\deg_v(i)+1}},
\label{eq:sevid}
\end{equation}
so that nodes with weaker neighborhood evidence receive larger penalties.

Finally, we define the proxy risk score used for conformal testing as
\begin{equation}
\hat{\kappa}(i)
=
S_{\mathrm{pred}}(i)+\beta\,S_{\mathrm{evid}}(i),
\label{eq:kappa_hat}
\end{equation}
where $\beta\ge 0$ balances instability and evidence.
Intuitively, $\hat{\kappa}(i)$ becomes large when predictions are sensitive to plausible perturbations or when the node is weakly supported by the constructed relations.

\paragraph{Conformal p-values.}
Given the proxy risk score $\hat{\kappa}(i)$, we test the risk hypothesis in \eqref{eq:hypothesis-risky} by constructing conformal p-values using the calibration set. Concretely, we partition nodes into three disjoint subsets,
$\Omega_{\mathrm{train}}$, $\Omega_{\mathrm{cal}}$, and $\Omega_{\mathrm{test}}$. 
We first define the calibration ``bad'' subset
\begin{equation}
\mathcal{B}=\{i\in\Omega_{\mathrm{cal}}:\ |y_i-\hat{y}(i)|\ge \delta\},
\label{eq:bad_set}
\end{equation}
which provides an empirical reference set for the null population.
A common concern for applying conformal prediction on graphs is that node dependencies may violate exchangeability. Recent studies show that conformal validity can still hold for node-level tasks on static graphs when the nonconformity score is invariant to permutations of the calibration/test nodes~\cite{zargarbashi2023conformal}. In our setting, calibration and test nodes are treated identically during graph construction: their features and edges are determined solely by observed covariates, with targets masked. Under this setting, we give the following exchangeability assumption between $\Omega_{\mathrm{cal}}$ and $\Omega_{\mathrm{test}}$.

\begin{assumption}[Exchangeability of calibration and test nodes]
\label{assump:exchangeability}
Conditioned on the observed covariates used and the trained model parameters, the collection of node-level examples in $\Omega_{\mathrm{cal}}\cup\Omega_{\mathrm{test}}$ is exchangeable under permutations of indices.
\end{assumption}

For each node $j$, we compute a conformal p-value \cite{jin2023selection} against $\mathcal{B}$ as
\begin{equation}
p_j
=
\frac{
1+\sum_{i\in\mathcal{B}}\mathbbm{1}\{\hat{\kappa}(i)<\hat{\kappa}(j)\}
+U_j\sum_{i\in\mathcal{B}}\mathbbm{1}\{\hat{\kappa}(i)=\hat{\kappa}(j)\}
}{
|\Omega_{\mathrm{cal}}|+1
},
\label{eq:conformal_p}
\end{equation}
where $U_j\sim\mathrm{Unif}[0,1]$ is used for random tie-breaking.
Intuitively, a smaller $p_j$ means that the current imputation has unusually low proxy risk compared with the calibration bad set $\mathcal{B}$, and is therefore less likely to be risky.

\begin{proposition}[Generalized validity under the risk null]
Following Jin and Cand\`es~\cite{jin2023selection}, we view $H_j$ as a random hypothesis.
Under Assumption~\ref{assump:exchangeability} and the validity conditions of
conformal selection for the proxy-risk score, for any test node $j$, the
$p$-value in~\eqref{eq:conformal_p} satisfies the generalized validity property:
\[
\mathbb{P}\!\left(p_j \le \alpha,\; H_j\right) \le \alpha,
\qquad
\forall \alpha \in [0,1].
\]
\end{proposition}

\paragraph{FDR-controlled selection via Benjamini--Hochberg.}
In clinical deployment, it is more meaningful to control the overall error burden among the imputed values that are actually presented to clinicians or downstream decision models.
We therefore target control of the false discovery rate (FDR) \cite{benjamini1995controlling}, defined as the expected proportion of truly risky imputations among those we accept for deployment.
To this end, we apply the Benjamini--Hochberg (BH) procedure \cite{jin2023selection, benjamini1995controlling} to the collection of p-values $\{p_j\}_{j\in\Omega_{\mathrm{test}}}$ at a target level $\alpha\in(0,1)$.
Let $p_{(1)}\le \cdots \le p_{(m)}$ be the sorted p-values over the $m=|\Omega_{\mathrm{test}}|$ test nodes, and let
\begin{equation}
k^\star=\max\Big\{k:\ p_{(k)}\le \frac{k}{m}\,\alpha\Big\},
\label{eq:bh_cutoff}
\end{equation}
with the convention that $k^\star=0$ if the set is empty. 
We then select the nodes
\begin{equation}
\mathcal{S}(\alpha)=\{j\in\Omega_{\mathrm{test}}:\ p_j\le p_{(k^\star)}\}.
\label{eq:bh_select}
\end{equation}
Since the null hypothesis $H_j$ corresponds to ``the imputation at $j$ is risky'', rejecting $H_j$ means selecting node $j$ as sufficiently reliable for release.
Therefore, $\mathcal{S}(\alpha)$ is the deployed subset with a user-specified FDR control level. 
\begin{proposition}[FDR control via BH]
\label{prop:fdr_bh}
Assume that the score-level conditional exchangeability and
no-ties conditions required by the exchangeable-data conformal
selection guarantee in \cite{jin2023selection} hold. Then applying BH at level $\alpha$ to $\{p_j\}_{j\in\Omega_{\mathrm{test}}}$ yields
\begin{equation}
\mathrm{FDR}
=
\mathbb{E}\!\left[
\frac{
\big|\{j\in\mathcal{S}(\alpha):\, |y_j-\hat{y}(j)|\ge \delta\}\big|
}{
\max\{|\mathcal{S}(\alpha)|,1\}
}
\right]
\le \alpha,
\end{equation}
i.e., among the selected imputations, the expected fraction of truly risky ones is controlled by $\alpha$.
\end{proposition}
The proof follows the conformal testing used in \cite{shen2025safer, jin2023selection}; for completeness we provide it in Appendix~\ref{app:fdr_proof}.

%% file: sec/4_experiments.tex
\section{Results}
In this section, we analyze four key aspects to demonstrate the effectiveness of \model{}: (i) overall imputation accuracy and reliability control of the proposed method compared with various baselines; (ii) the performance of the pure event-graph imputer without FDR control; (iii) an ablation study that quantifies the impact of major design components; and (iv) a detailed analysis of FDR-controlled selection.

\begin{table*}[t]
\centering

\caption{Overall performance comparison. Best results are in \textbf{bold} and second-best are \underline{underlined}.}
\vspace{-5pt}
\label{tab:classic_method}
\begin{tabular*}{0.95\textwidth}{@{\extracolsep{\fill}} ll ccc ccc ccc}
\toprule
\multirow{2}{*}{\textbf{Method}} & \textbf{Dataset} &
\multicolumn{3}{c}{\textbf{Mayo Clinic}} &
\multicolumn{3}{c}{\textbf{MIMIC-III}} &
\multicolumn{3}{c}{\textbf{MIMIC-IV}} \\
\cmidrule(lr){3-5}\cmidrule(lr){6-8}\cmidrule(lr){9-11}
& \textbf{Metric}
& MAE $\downarrow$ & RMSE $\downarrow$& Prec. $\uparrow$
& MAE $\downarrow$& RMSE $\downarrow$& Prec. $\uparrow$
& MAE $\downarrow$& RMSE $\downarrow$& Prec. $\uparrow$ \\
\midrule
\midrule

\multirow{5}{*}{\textbf{\textit{Statistical}}}
& Mean        & 0.4311 & 0.5086 & 0.8333 & 1.2528 & 1.4844 & 0.3970 & 1.1601 & 1.4991 & 0.5044 \\
& KNN         & 0.4983 & 0.5776 & 0.5000 & 1.2758 & 1.5844 & 0.4879 & 1.2962 & 1.6638 & 0.4557 \\
& MICE        & 0.3889 & 0.4589 & 0.8333 & 1.0382& 1.2747 & 0.5394 & 0.9960 & 1.3906 & 0.6346 \\
& Missforest  & 0.3419 & 0.4416 & 0.8333  & 1.1409 & 1.3614 & 0.4697 & 0.9864 & \underline{1.3351} & 0.6185 \\
& Hyperimpute & 0.3833 & 0.4819 & 0.8333 & 0.9844 & \underline{1.2409} & 0.5758 & 0.9872 & 1.3693 & 0.6185 \\

\midrule

\multirow{6}{*}{\textbf{\textit{Deep Learning}}}

& GAIN        & 0.3635 & 0.4360 & 0.8333     & 1.0613 & 1.3510 & 0.5394 & 1.0148 & 1.4378 & 0.6275 \\
& GRAPE       & \underline{0.2494} & 0.3688 & 0.8333 & \underline{0.9724} & 1.2526 & \underline{0.6333} & 1.0057 & 1.6450& \underline{0.6487}  \\
& TDM         & 0.4191 & 0.4832 & 0.6667     & 0.9934 & 1.2543 & 0.6091 & 1.1313 & 1.4749 & 0.5143 \\
& MIWAE       & 0.5040 & 0.6054   & 0.6667 & 1.1782 & 1.5031 & 0.5182 & 1.0124 & 1.4118 & 0.6232 \\
& Remasker    & 0.5250 & 0.6475 & 0.8333     & 1.1112 & 1.3463 & 0.5061 & 1.2324 & 1.5260 & 0.4431 \\
& DIFFIMPUTER & 0.3060 & 0.4026 & \underline{1.0000}     & 1.0771 & 1.3122 & 0.5212 & \underline{0.9791} & \textbf{1.2305} & 0.6400 \\

\midrule

\multirow{2}{*}{\textbf{\textit{Sequential}}}
& LSTM        & 0.6620 & 0.7771 & 0.8333 & 1.1633 & 1.4025 & 0.4455 & 0.9934 & 1.4242 & 0.6293 \\
& TRANS       & 0.2649 & \underline{0.3129} & \underline{1.0000} & 2.1025 & 2.4062 & 0.1939 & 1.1627 & 1.9186 & 0.6052 \\

\midrule

& \textbf{SafeImpute}
& \textbf{0.2464} & \textbf{0.3125} & \textbf{1.0000}
& \textbf{0.9662} & \textbf{1.2351} & \textbf{0.6333}
&\textbf{0.9715} & 1.3820 & \textbf{0.6555} \\

\bottomrule
\end{tabular*}
\vspace{-5pt}
\end{table*}

\subsection{Experimental Settings}

\subsubsection{Datasets}
We evaluate \model~ on HbA1c (A1c) imputation for diabetes patients, where HbA1c is a key indicator for long-term glycemic control and diabetes management, using three datasets: Mayo Clinic, MIMIC-III \cite{johnson2016mimic}, and MIMIC-IV \cite{johnson2023mimic}. Mayo Clinic dataset contains real-world records collected in 2022--2023 from patients diagnosed with type~1 diabetes, where each patient has multiple visits; for each visit, we extract \{A1c, glucose, cholesterol, gender\} as features, and the lab variables exhibit naturally occurring missingness because tests are not ordered at every visit. To create a test set for evaluation, we additionally mask 10\% of observed A1c values as held-out targets. For MIMIC-III and MIMIC-IV, we construct diabetes cohorts by filtering patients with diabetes diagnoses and treat each A1c measurement as an anchor visit; for each anchor, we retrieve the most recent glucose and cholesterol measurements recorded at or before the A1c timestamp (within 7 days for glucose and 90 days for cholesterol), and use gender as the demographic feature, yielding irregular longitudinal event sequences with naturally occurring missingness across patients. We randomly mask 40\% of observed A1c values as held-out targets. Apart from the held-out test nodes, we split the remaining nodes into training/validation/calibration sets with a 70\%/15\%/15\% ratio. Statistics of processed datasets and detailed analysis of Mayo Clinic data are provided in the appendix \ref{app:data}. Definitions of the evaluation metrics are provided in Appendix~\ref{app:exp-details}.

\subsubsection{Baselines}
Existing imputation baselines generally do not provide native uncertainty estimation or selective-release mechanisms. 
Since prior work on uncertainty estimation for imputation~\cite{hossain2025beyond} suggests that methods such as MICE~\cite{little2019statistical} and MIWAE~\cite{mattei2019miwae} can reflect both calibration and accuracy, we include a diverse set of baselines spanning statistical, deep learning, and sequential methods.
Statistical methods: Mean imputes each feature by the training-set mean. KNN imputes from the k nearest samples based on similarity over observed dimensions. MICE~\cite{little2019statistical} iteratively fits feature-wise conditional regressors to update missing entries. MissForest~\cite{stekhoven2012missforest} performs iterative imputation using random-forest predictors. HyperImpute \cite{jarrett2022hyperimpute} selects a strong imputer via automated model and hyperparameter search.
Deep learning methods: GAIN~\cite{gain} uses adversarial training to generate realistic imputations. GRAPE~\cite{grape} leverages graph neural message passing to propagate information for missing values. MIWAE~\cite{mattei2019miwae} performs imputation under a variational generative framework. TDM~\cite{zhao2023transformed} models complex tabular dependencies with a deep imputation architecture. ReMasker~\cite{du2024remasker} learns masked reconstruction with a denoising objective. DiffImputer~\cite{zhang2025diffputer} uses diffusion-based denoising to generate completed samples.
Sequential methods: LSTM~\cite{hochreiter1997long} imputes using recurrent hidden states over irregular event sequences. TRANS~\cite{yang2023transformehr} is a transformer-based sequential model for electronic health records data prediction; we adapt it for imputation by casting missing-value recovery as a prediction problem over masked targets. We evaluate all methods on the same set of predictions selected by the proxy risk score and report each method's empirical Precision based on its own imputation outputs.

\subsubsection{Implementation} 

We choose the target FDR level $\alpha$ according to the empirical FDR--acceptance trade-off on each dataset. Specifically, we set $\alpha{=}0.15$ for Mayo Clinic and $\alpha{=}0.35$ for MIMIC-III/IV. Very strict target levels can yield an extremely small or empty accepted set, making the selective results unstable and less practically useful. We set the clinically interpretable threshold $\delta$ to $0.6$ percentage points for Mayo Clinic and $1.0$ percentage point for MIMIC-III/IV. For all datasets, we use an auxiliary loss weight $\lambda{=}0.1$ and set $\beta{=}0.1$ when computing the proxy risk score. All experiments are run on a single NVIDIA H200 GPU. For baselines, we use the default PyPI implementation for MICE; for HyperImpute and MissForest, we use their official packages; and for KNN, we set $k{=}10$. For other baselines, we adopt the original implementations and follow the hyperparameter recommendations in the corresponding papers, with minor tuning for each dataset, and report the best-performing configuration. Each experiment is repeated with five random seeds, and we report the mean performance.

\subsection{Overall Performance}

Table \ref{tab:classic_method} shows the overall performance of \model{} against a broad set of baselines on Mayo Clinic, MIMIC-III, and MIMIC-IV. Overall, \model{} is consistently top-performing across datasets, demonstrating strong performance with FDR control and robustness under irregular and sparse missing data. Beyond the overall gains, comparing baseline behaviors to our method yields several findings. (i) Among non-neural baselines that are still commonly used in practice, MissForest and HyperImpute are relatively strong competitors, but they become less competitive in more challenging settings (e.g., when missingness is severe); notably, after applying our FDR-controlled selection, these simple methods also improve compared to their uncontrolled results (Table \ref{tab:event_graph}), suggesting that risk-aware selection can be a broadly useful add-on for simple imputers. (ii) LSTM becomes less competitive, indicating that when sufficient temporal context is not available, classic time-series models remain sensitive to irregular long gaps and sparse supervision. (iii) GRAPE, as a purely graph-based method, performs strongly, reinforcing the effectiveness of capturing intra-patient and inter-patient relations via the graph. (iv) TRANS achieves comparable performance on Mayo Clinic, implying that combining a temporal Transformer with graph modeling can be effective when sequences are short and irregular; at the same time, our comparison suggests that high-quality error control often requires method-aware design to support more reliable selection.

\begin{table}[t]
\centering
\small
\setlength{\tabcolsep}{3.0pt} 
\caption{Event graph imputation without FDR control.}
\vspace{-5pt}
\label{tab:event_graph}
\renewcommand{\arraystretch}{1.05} 
\begin{tabular*}{\columnwidth}{@{\extracolsep{\fill}} l cc cc cc}
\toprule
\multirow{2}{*}{\textbf{Method}} &
\multicolumn{2}{c}{\textbf{Mayo Clinic}} &
\multicolumn{2}{c}{\textbf{MIMIC-III}} &
\multicolumn{2}{c}{\textbf{MIMIC-IV}} \\
\cmidrule(lr){2-3}\cmidrule(lr){4-5}\cmidrule(lr){6-7}
& MAE & RMSE & MAE & RMSE & MAE & RMSE \\
\midrule

Mean        & 0.6612 & 0.8198 & 1.3455 & 1.8495 & 1.3016 & 1.7386 \\
KNN         & 0.7479 & 0.9239 & 1.3785 & 1.9068 & 1.1507 & 1.6223 \\
MICE        & 0.6369 & 0.8045 & 1.1787 & 1.6490 & 1.1056 & 1.5231 \\
Missforest  & 0.6300 & 0.8010 & 1.1523 & \underline{1.6148} &   1.1233     & 1.5490       \\
Hyperimpute & 0.6640 & 0.8263 & 1.1664 & 1.6396 &  1.1441      & 1.5725 \\

GAIN        &  0.7200      &0.8564        & 1.4251 & 1.8650 & 1.2270 & 1.7497 \\
GRAPE       &   \underline{0.5655}     &   0.7743     & 1.2072 & 1.8092 & 1.1509 & 1.8945 \\
TDM         &   0.6213     & 0.7946       &   1.1816     &   1.7180     &   1.2578     & 1.7052       \\
Remasker    &     0.6654   &  0.8329      &   1.2120     & 1.7055      &    1.3179    & 1.6859      \\
DIFFIMPUTER &   0.6227     &   0.7988     & \underline{1.1411} & 1.7129 & \underline{1.0467} & \textbf{1.4541} \\

LSTM        & 0.9289 & 1.1666 & 1.3377 & 1.8428 & 1.1449 & 1.7672 \\
TRANS        & 0.5991 & \underline{0.7696} & 1.5403 & 1.8801 & 1.4214 & 2.2184 \\
\midrule

\textbf{SafeImpute}    & \textbf{0.4142} & \textbf{0.5253} & \textbf{1.0621} & \textbf{1.6002} & \textbf{0.9642 }& \underline{1.5181} \\
\bottomrule
\end{tabular*}
\vspace{-10pt}
\end{table}

\subsection{Event Graph Evaluation}

We then evaluate the pure imputation performance on the event graph \emph{without} any error control. In this setting, we construct edges only using the proposed temporal and trend-aware value criteria, and we do not add extra value-only edges when no eligible trend-based neighbor exists, to avoid noisy connections. Table~\ref{tab:event_graph} reports results on Mayo Clinic, MIMIC-III, and MIMIC-IV. The proposed event-graph imputer achieves the best or second best performance across all three datasets, highlighting the benefit of event-graph construction and a two-relation GNN that combines intra-patient temporal continuity with inter-patient value-based similarity under irregular and sparse visit patterns. Sequential baselines are less competitive, suggesting that irregular visits and sparse observations limit sequence-only modeling, while classical imputers remain behind because feature-conditional modeling cannot fully capture latent relational signals under severe missingness. Additionally, the pure event-graph imputer sometimes slightly outperforms the FDR-controlled version because the latter adds value-only fallback edges to support reliability estimation, which may introduce noisy cross-patient connections. This reflects a small accuracy and reliability trade-off for selective release.

\vspace{-5pt}
\subsection{Ablation Study}
\label{sec:ablation}
We conduct ablation studies on Mayo Clinic and MIMIC-III to assess both the event-graph imputer and the reliability-control module. Removing temporal edges or trend-aware value edges degrades performance, showing that intra-patient temporal continuity and inter-patient clinical similarity provide complementary signals. Removing the adaptive fusion gate causes a substantial drop, highlighting the importance of dynamically balancing the two relations. The auxiliary reconstruction loss improves performance on Mayo Clinic and serves as an additional regularizer for learning event representations from sparse non-target labs. 

We also ablate the proxy risk score for FDR-controlled selection. Using prediction instability alone leads to much higher FDR than the full proxy score, increasing FDR from 0.0000 to 0.3575 on Mayo Clinic and from 0.3667 to 0.6304 on MIMIC-III. This confirms that the evidence penalty is important for reliable release under sparse relational support.

\begin{table}[t]
\centering
\small
\caption{Ablation study of the event-graph imputer without FDR-controlled selection.}
\vspace{-5pt}
\label{tab:ablation_imputer}
\begin{tabular}{lcccc}
\toprule
\multirow{2}{*}{\textbf{Variant}} &
\multicolumn{2}{c}{\textbf{Mayo}} &
\multicolumn{2}{c}{\textbf{MIMIC-III}} \\
\cmidrule(lr){2-3}\cmidrule(lr){4-5}
& MAE & RMSE & MAE & RMSE \\
\midrule
w/o temporal edges 
& 0.9669 & 1.2231 & 1.2889 & 1.8091 \\
w/o trend-aware value edges 
& \underline{0.4213} & \textbf{0.5141} & 1.0902 & 1.7683 \\
w/o adaptive fusion gate 
& 0.9773 & 1.2373 & 1.1192 & 1.7312 \\
w/o auxiliary reconstruction 
& 0.4434 & 0.5485 & \textbf{1.0355} & \textbf{1.5724} \\
SafeImpute 
& \textbf{0.4142} & \underline{0.5253} & \underline{1.0621} & \underline{1.6002} \\
\bottomrule
\end{tabular}
\vspace{-5pt}
\end{table}

\begin{table}[t]
\centering
\small
\caption{Ablation study of the proxy risk score under FDR-controlled selection.}
\vspace{-5pt}
\label{tab:ablation_proxy}
\begin{tabular}{llccc}
\toprule
\textbf{Dataset} & \textbf{Variant} & \textbf{MAE} & \textbf{RMSE} & \textbf{FDR} \\
\midrule
\multirow{2}{*}{Mayo}
& Instability-only score & 0.3150 & 0.3675 & 0.3575 \\
& Proxy risk score & \textbf{0.2464} & \textbf{0.3125} & \textbf{0.0000} \\
\midrule
\multirow{2}{*}{MIMIC-III}
& Instability-only score & 1.1468 & 1.6246 & 0.6304 \\
&  Proxy risk score & \textbf{0.9662} & \textbf{1.2351} & \textbf{0.3667} \\
\bottomrule
\end{tabular}
\vspace{-5pt}
\end{table}

\begin{figure*}[ht]
    \centering
    \includegraphics[width=0.24\textwidth]{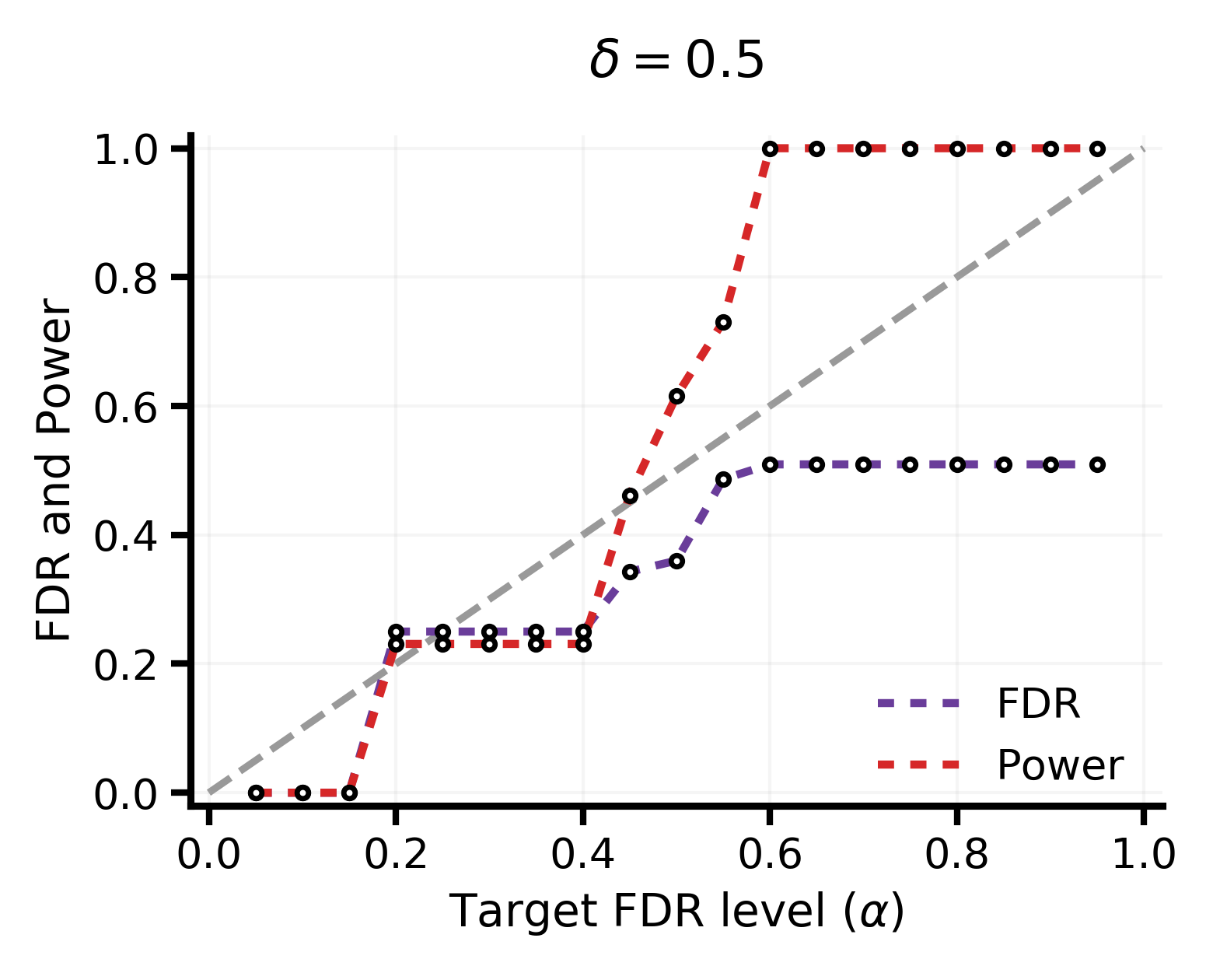}
    \includegraphics[width=0.24\textwidth]{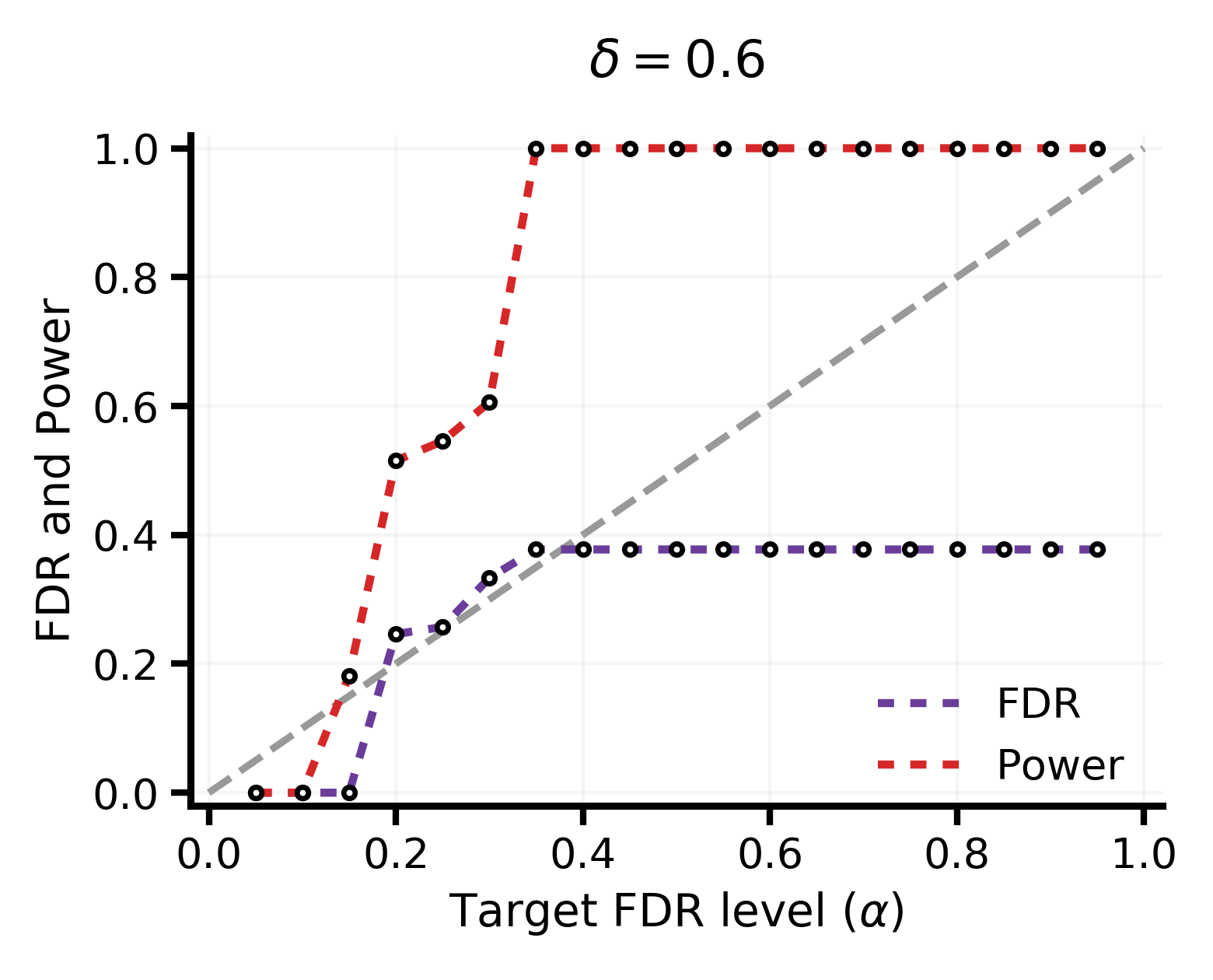}
    \includegraphics[width=0.24\textwidth]{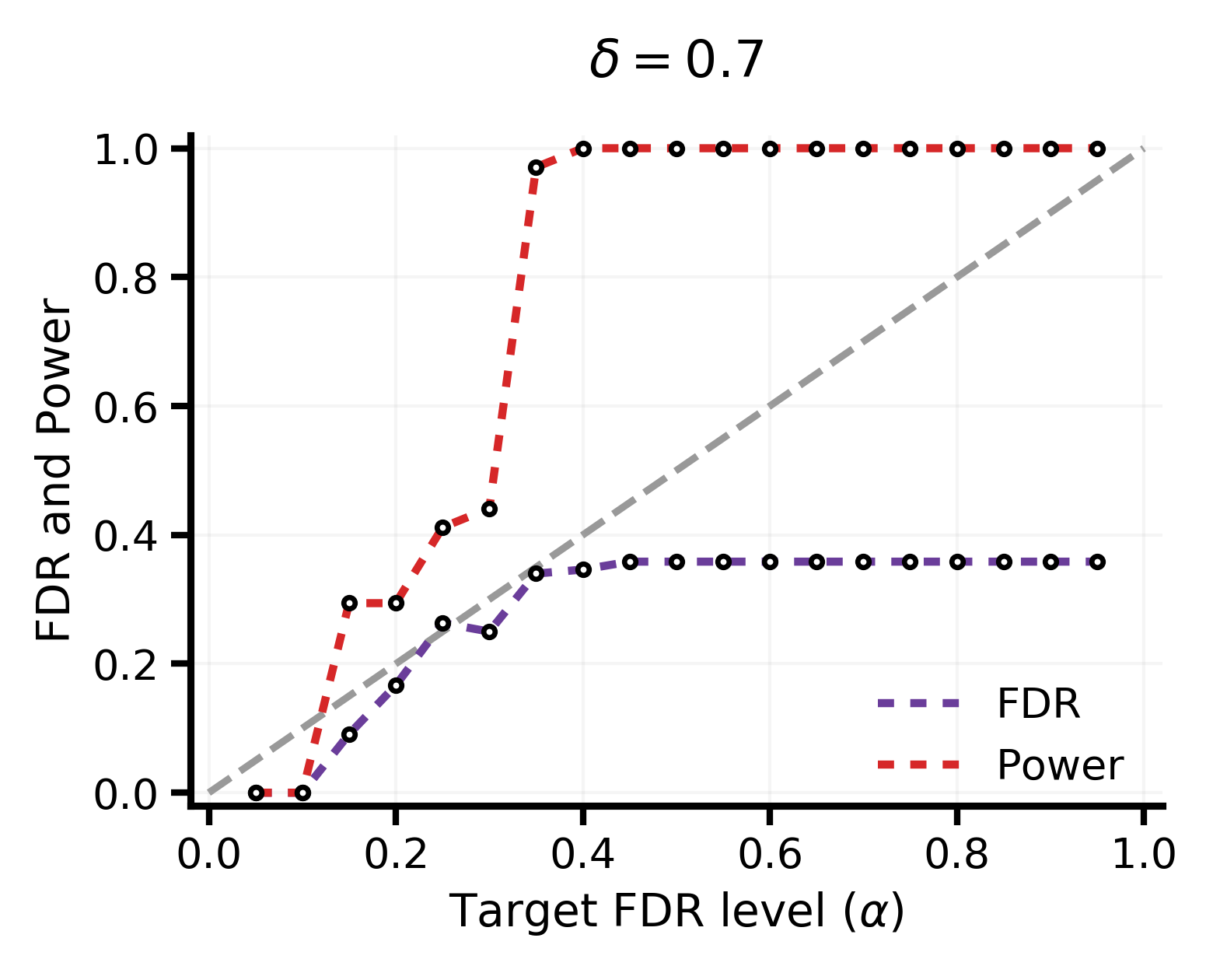}
    \includegraphics[width=0.24\textwidth]{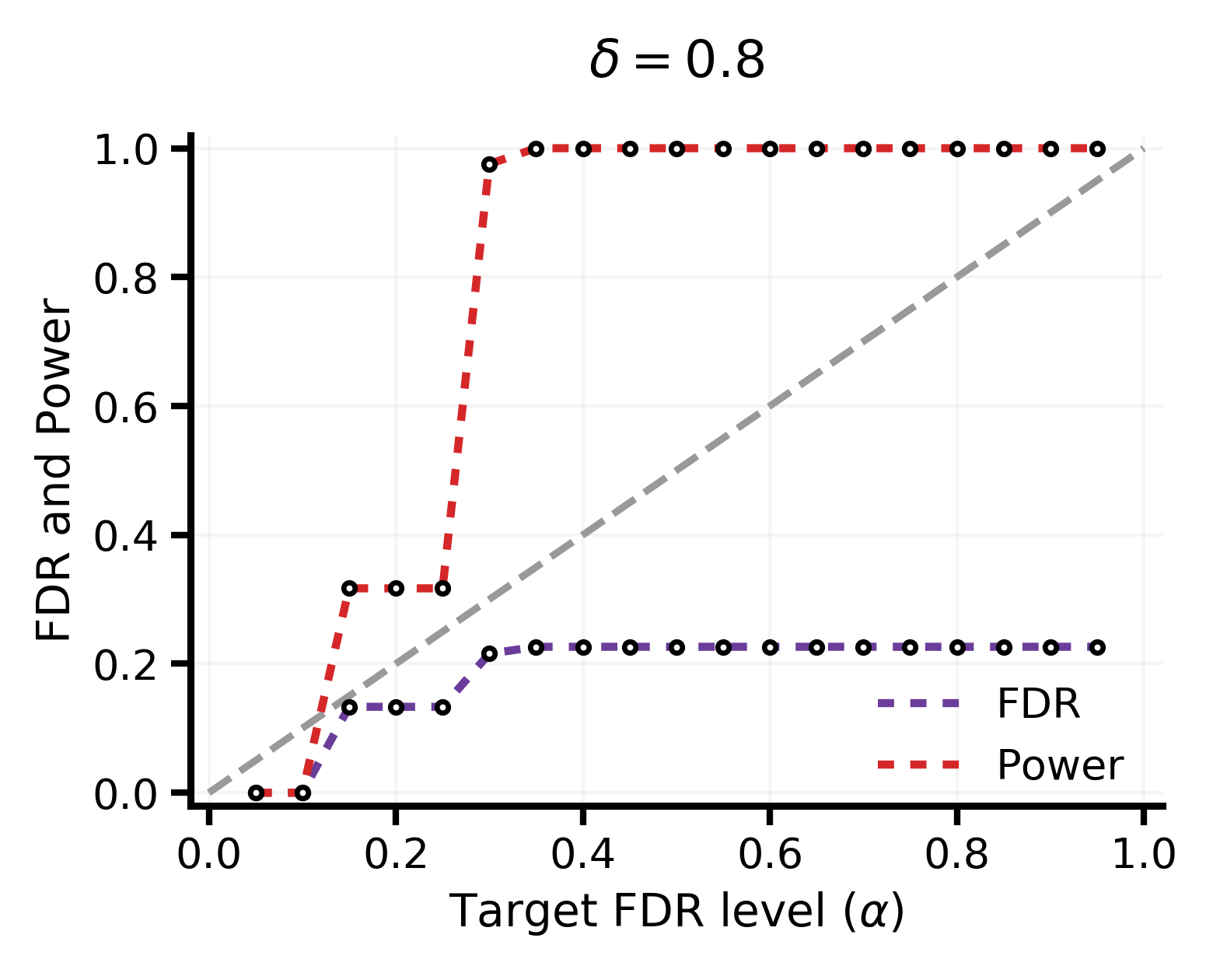}

\vspace{-10pt}
    \caption{FDR--power trade-off of conformal selection on Mayo Clinic dataset.
Each panel fixes the clinical tolerance $\delta$ and sweeps the target FDR level $\alpha$.
The dashed gray line indicates perfect calibration (FDR$=\alpha$).\label{fig:mayo_fdr_power_delta}
}
    \vspace{-8pt}
\end{figure*}

\vspace{-5pt}
\subsection{A Close Look at FDR Control}

\paragraph{FDR and Power.}
Recall that our test-time null hypothesis is $H_j:\ |y_j-\hat{y}(j)|\ge \delta$, i.e., node $j$ has a risky imputation.
Our goal is to reject these risky-imputation nulls and retain nodes whose imputations are sufficiently reliable for release.
Figure~\ref{fig:mayo_fdr_power_delta} sweeps the BH target level $\alpha$ and reports the FDR among the selected set, defined as the fraction of selected nodes that are in fact risk points (with $|y_j-\hat{y}(j)|\ge\delta$).
We also report power, measured as the fraction of non-risk nodes (with $|y_j-\hat{y}(j)|<\delta$) that are successfully retained.
Across $\delta\in\{0.5,0.6,0.7,0.8\}$, the achieved FDR closely follows the target, indicating conservative control.
As $\alpha$ increases, more hypotheses are rejected and more nodes are selected, leading to higher power and eventual saturation when the BH cutoff becomes non-restrictive.
Larger tolerances $\delta$ make the risky-imputation null less likely to hold, resulting in fewer true risk points; consequently, the retained set attains higher power and lower FDR, consistent with the intended screening behavior. FDR curves on MIMIC-III and MIMIC-IV are provided in Appendix~\ref{app:fdr_power_mimic}.

\begin{figure}[h]
    \centering

    \begin{subfigure}[t]{0.45\linewidth}
        \centering
        \includegraphics[width=\linewidth]{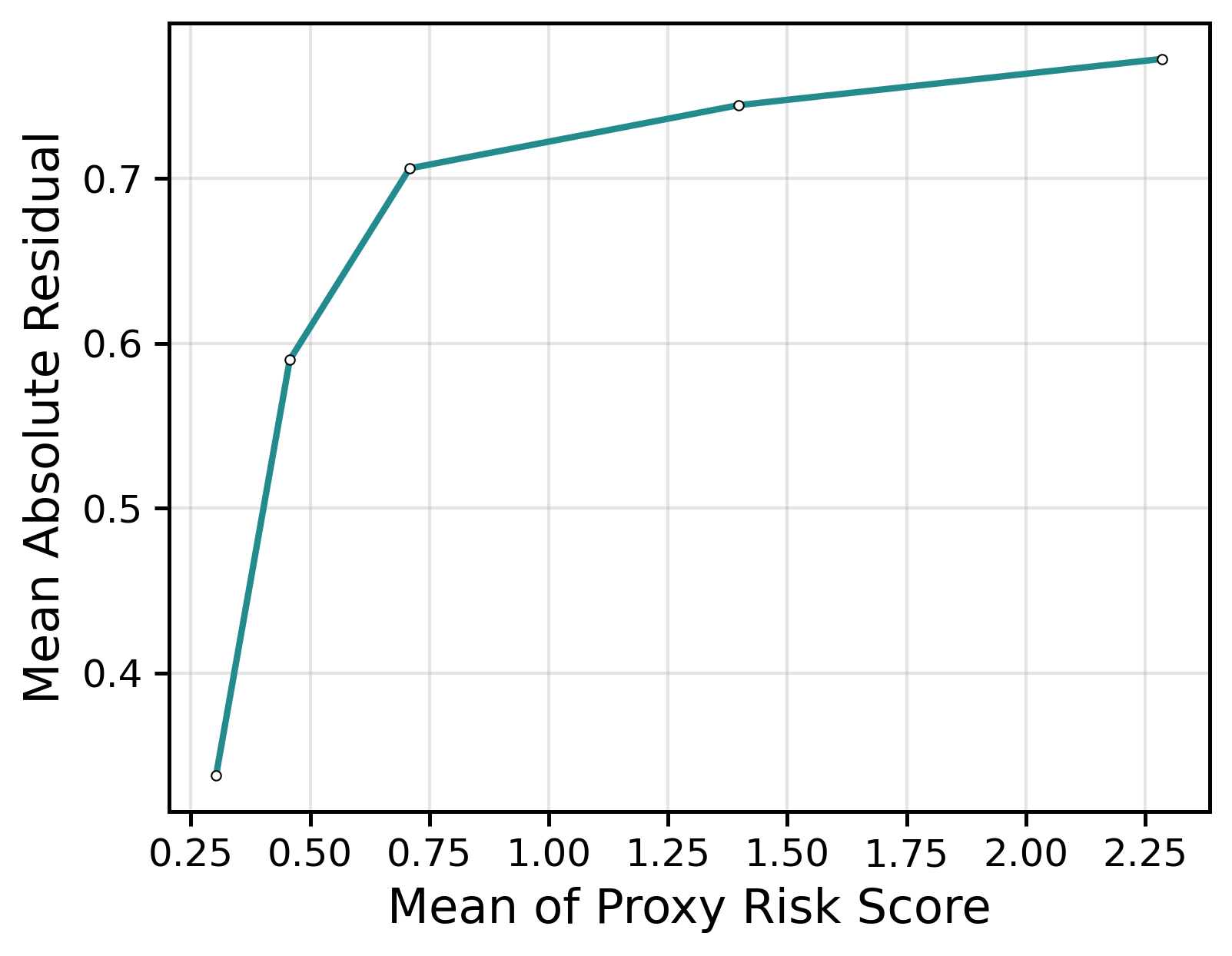}
        \caption{Mayo Clinic}
        \label{fig:mayo_fdr_power_d05}
    \end{subfigure}
    \hfill
    \begin{subfigure}[t]{0.45\linewidth}
        \centering
        \includegraphics[width=\linewidth]{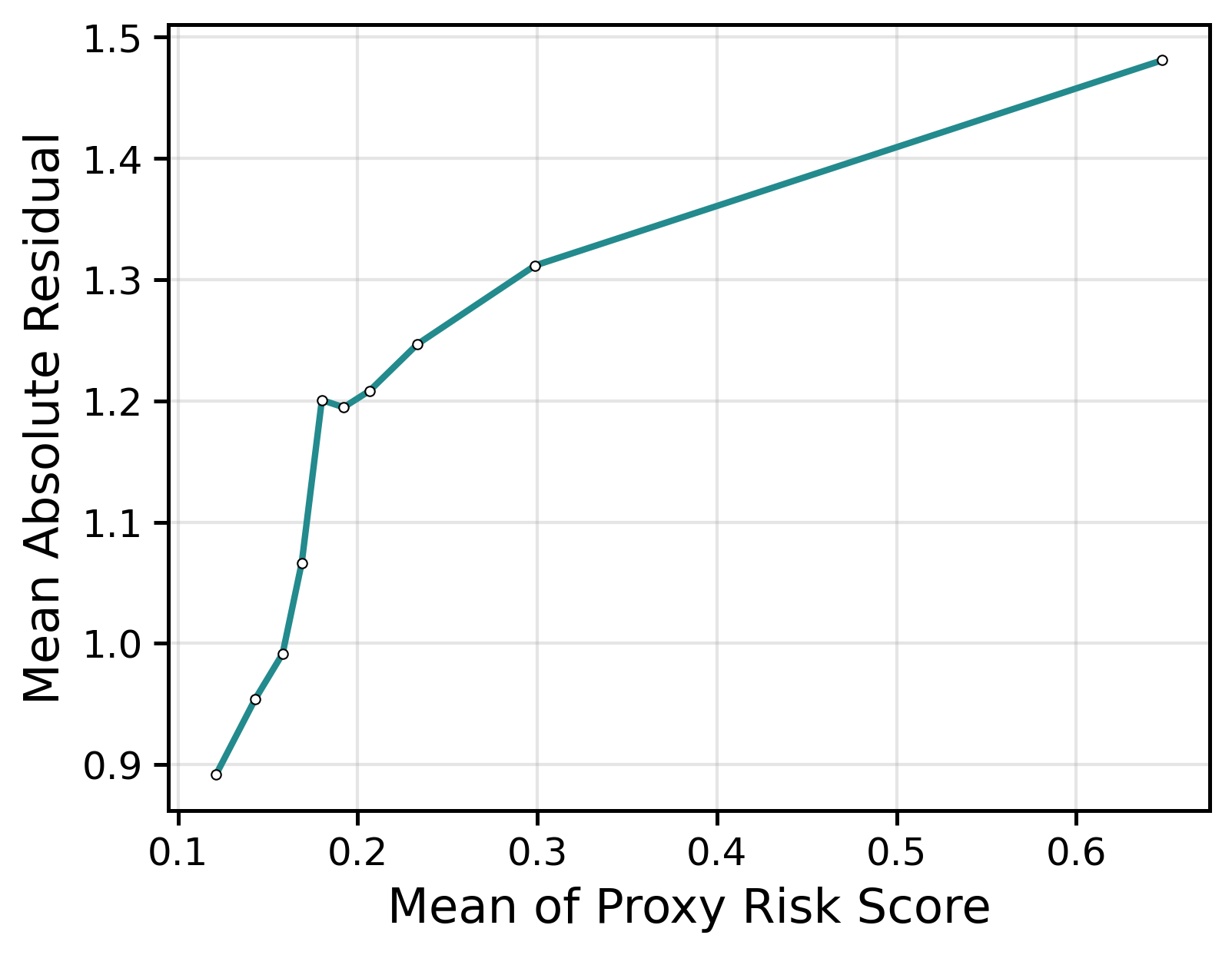}
        \caption{MIMIC-III}
        \label{fig:mayo_fdr_power_d06}
    \end{subfigure}
\vspace{-5pt}
    \caption{Alignment between the proxy risk score $\hat{\kappa}$ and the real imputation error.}

    \vspace{-10pt}
    \label{fig:risk_score_mean}
\end{figure}

\paragraph{Selective Release Summary}

Table~\ref{tab:selective_release_summary_app} summarizes the selective release results under the chosen operating point for each dataset. 
The retained subset is selected by the pre-specified proxy risk score, conformal calibration, and the Benjamini--Hochberg procedure, without access to test labels during selection. 
The results show that \model{} retains a nontrivial fraction of imputations under the target FDR constraint, rather than achieving error control by selecting only a very small subset. 
The acceptance fraction reflects the coverage of released imputations under the target FDR constraint, while power measures how many truly reliable imputations are successfully retained. 
Across datasets, the selected operating points illustrate the trade-off between empirical FDR and retained coverage: stricter selection yields lower FDR but smaller acceptance fractions, while more permissive target levels retain more imputations at the cost of higher empirical FDR.

\begin{table}[t]
\centering
\caption{Selective release results under the chosen target FDR levels. 
FDR measures the fraction of clinically unacceptable errors among retained imputations, while power measures the fraction of truly reliable imputations that are retained.}
\vspace{-5pt}
\label{tab:selective_release_summary_app}
\resizebox{0.98\linewidth}{!}{
\begin{tabular}{lcccc}
\toprule
Dataset & FDR $\downarrow$ & Power $\uparrow$ & Acceptance Fraction $\uparrow$ & Precision $\uparrow$ \\
\midrule
Mayo Clinic & 0.000 & 0.1818 & 0.1132 & 1.000 \\
MIMIC-III   & 0.3667 & 0.0982 & 0.0883 & 0.6333 \\
MIMIC-IV    & 0.3437 & 0.2163 & 0.1822 & 0.6563 \\
\bottomrule
\end{tabular}
}
\vspace{-10pt}
\end{table}

\paragraph{Quality of the Proxy Risk Score.}
Our FDR-controlled selection relies on conformal p-values constructed from the proxy risk score $\hat{\kappa}$, which in turn determines the BH rejection set.
A well-behaved proxy should ideally exhibit a monotonic relationship with the realized imputation error: nodes with larger $\hat{\kappa}$ should, on average, have larger realized errors $|y-\hat{y}|$.
Such alignment is crucial because it makes the ranking induced by $\hat{\kappa}$ informative for separating high-error nodes from low-error ones. Here, we evaluate this property on both the Mayo Clinic and MIMIC-III datasets.
For each dataset, we partition evaluation nodes into quantile groups (5 for Mayo Clinic and 10 for MIMIC-III) according to $\hat{\kappa}$ and plot the mean absolute error within each group against the mean proxy score.
Figure~\ref{fig:risk_score_mean} shows a clear increasing trend on both datasets: groups with higher average $\hat{\kappa}$ consistently have higher mean absolute residuals.
This indicates that $\hat{\kappa}$ is a meaningful surrogate of the unobserved error and is suitable for conformal selection when test labels are unavailable.

%% file: sec/5_related_work.tex
\section{Related Work}
\subsection{Data Imputation}
Missing data is a pervasive challenge in many applications \cite{he-etal-2025-llm}. Classical methods,\textit{ e.g.}, simple heuristics (mean/median substitution) and machine learning methods such as KNN imputation~\cite{troyanskaya2001missing} 
offer straightforward solutions but rely on restrictive assumptions and ignore feature relationships. Deep learning approaches leverage various techniques and training paradigms \cite{sankar2022self, infomotif, 10.1145/3583780.3615039, wang2025multi} (e.g., GANs, GNNs, and diffusion models). Representative examples include GAIN~\cite{gain}, which trains a generator to impute missing entries conditioned on observed entries while a discriminator distinguishes observed entries from imputed ones; GRAPE~\cite{grape}, which represents tabular data \cite{zou2026rag} as a bipartite graph with observed values treated as edges and formulates imputation as an edge-level prediction task; and DiffPuter~\cite{zhang2025diffputer}, which combines diffusion models with an EM-style training procedure to learn the joint distribution of complete data and perform conditional sampling for missing entries. Autoencoder-based methods~\cite{vincent2008extracting, gondara2018mida}, such as ReMasker~\cite{du2024remasker}, adapt masked autoencoding to tabular imputation by masking entries and training a deep learning-based model to reconstruct entries from context. Data imputation is particularly important in healthcare settings, where missing data is common due to irregular measurements and clinical workflows. Many of these ideas have also been adapted to clinical records. For example, 
GRU-D~\cite{che2018recurrent} and BRITS~\cite{cao2018brits} leverage RNNs to model temporal dynamics. Attention-based models such as SAITS~\cite{du2023saits} use self-attention with self-supervised masking to better capture long-range dependencies of clinical variables. GRIN~\cite{cini2021filling} exploits GNNs \cite{10.1145/3404835.3462857, zheng2025pyg} and performs message passing across variables and time. 
More recently, diffusion-based approaches such as CSDI~\cite{tashiro2021csdiconditionalscorebaseddiffusion} perform conditional denoising to iteratively refine missing values and have shown strong performance on ICU benchmarks. Although these methods are effective, they primarily optimize the reconstruction error and overlook the reliability of imputations.
\vspace{-5pt}
\subsection{Uncertainty Quantification in Healthcare Models} 

Given the high stakes of the domain, modern healthcare machine learning models increasingly incorporate uncertainty quantification (UQ) techniques to improve trustworthiness and safety~\cite{jabbour2025limitsselectiveaiprediction,CAMPAGNER2025106014}.  
Bayesian approaches \cite{blundell2015weight,lopez2025uncertainty} place probabilistic priors on model parameters to capture epistemic uncertainty and have been used in medical applications \cite{mehrtash2020confidence, zhao2022efficient} to indicate low-confidence predictions. For example, Qiu et al. \cite{qiu2019modeling} applied a Bayesian neural network to clinical records and found that instances with high predictive uncertainty were harmful to overall model performance. Related approximations include Monte Carlo dropout~\cite{hammersley2013monte}, which performs stochastic inference by enabling dropout at test time; TrUE-Net~\cite{jo2025uncertainty} applies this strategy to AD genomic variant classification. 
Deep ensembles are another empirical way to improve predictive performance and quantify the uncertainty via inter-model variability. DEGU \cite{zhou2026uncertainty} integrates ensemble learning to improve the robustness and explainability of genomic DNNs. Another complementary direction is conformal prediction (CP), which offers distribution-free uncertainty quantification with formal coverage guarantees through calibrated prediction sets. Building on CP, conformal selection \cite{jin2023selection} accepts predictions only when the conformal prediction set is sufficiently informative, yielding an explicit and risk-controllable mechanism for accepting a prediction \cite{vazquez2022conformal}. This paradigm has been increasingly adopted in recent clinical studies~\cite{jin2023selection,angelopoulos2024conformal, genomic_prediction}. SAFER~\cite{shen2025safer} uses conformal inference to provide statistical guarantees while filtering out uncertain treatment recommendations. 
Together, these advances motivate the integration of uncertainty quantification to support more reliable and risk-aware clinical AI.

%% file: sec/6_conclusion.tex
\vspace{-10pt}
\section{Conclusion}
We propose \model{}, a framework for reliable clinical data imputation that models irregular and sparse longitudinal records by constructing an event graph capturing both intra-patient temporal structure and inter-patient clinical similarity. 
For reliability control, \model{} calibrates a proxy risk score into conformal p-values and applies the Benjamini--Hochberg procedure to control the false discovery rate of clinically unacceptable imputation errors. Extensive experiments demonstrate that \model{} achieves strong imputation performance while providing meaningful reliability guarantees, outperforming selected baselines in both standard and FDR-controlled imputation evaluation. Future work includes: (i) incorporating multi-modal information and multi-source clinical data; and (ii) moving beyond imputation metrics to evaluate the effect of reliable imputations on downstream tasks, such as risk prediction and treatment recommendation.

\vspace{-5pt}
\section*{Ethical Use of Data}
To protect patient privacy, all data were anonymized during collection and processing. The use of this dataset has been reviewed and approved by the Institutional Review Board (IRB) to ensure compliance with ethical standards and participant protections. We strictly adhere to relevant laws and regulations regarding data use and protection to ensure the legal and compliant use of the data.
\vspace{-8pt}
\section*{Acknowledgments}
This work is supported by National Science Foundation under Award No. IIS-2117902. The views and conclusions are those of the authors and should not be interpreted as representing the official policies of the funding agencies or the government.

%% file: sec/appendix_1.tex
\section{Proof of FDR Control}
\label{app:fdr_proof}

\paragraph{Goal.}
Let $\kappa(i)=|y_i-\hat{y}(i)|$ be the true imputation error on labeled nodes, and fix a clinical tolerance $\delta>0$. For each test node $j\in\Omega_{\mathrm{test}}$ we test the null
\begin{equation}
H_j:\ \kappa(j)\ge \delta,
\end{equation}
i.e., the imputation at $j$ is \emph{risky}.

Let $\hat{\kappa}(x)$ be our label-free proxy risk score in~\eqref{eq:kappa_hat}. To apply the selective conformal testing argument in \cite{jin2023selection}, we define a nonconformity score using the binary risk label $\mathbbm{1}\{\kappa(i)\ge \delta\}$. Let $M>\sup_i \hat{\kappa}(i)$ (finite since $\hat{\kappa}$ is bounded in practice). Define, for any node $i$ and candidate risk label $b\in\{0,1\}$,
\begin{equation}
J(x,b)
=
\hat{\kappa}(x)
+
2M\cdot \mathbbm{1}\{b=0\}.
\label{eq:J_def}
\end{equation}
Thus, if the node is \emph{risky} ($b=1$), then $J(x,b)=\hat{\kappa}(x)$; if it is not risky ($b=0$), then
$J(x,0)=\hat{\kappa}(x)+2M$. Equivalently, $J$ is monotone in the reliability label $1-b$, which is the orientation used in the conformal selection construction
of \cite{jin2023selection}.

For each calibration node $i\in\Omega_{\mathrm{cal}}$, define its observed risk label
\begin{equation}
B_i=\mathbbm{1}\{\kappa(i)\ge \delta\},
\end{equation}
and for each test node $j\in\Omega_{\mathrm{test}}$, under the null $H_j$ we have $B_j=1$.
Let
\begin{equation}
J_i = J(i,B_i)\quad (i\in\Omega_{\mathrm{cal}}),
\qquad
\widehat{J}_j = J(j,1)\quad (j\in\Omega_{\mathrm{test}}),
\end{equation}
i.e., we evaluate the test score at the null label $b=1$.

Since $M>\sup_i\hat{\kappa}(i)$, for any calibration node with
$B_i=0$,
\begin{equation}
J_i=\hat{\kappa}(i)+2M>\widehat{J}_j.
\end{equation}
Therefore, such nodes do not contribute to the lower-tail rank. Only calibration nodes with $B_i=1$, i.e., the bad calibration subset used in the main text, contribute to the conformal p-value.

\paragraph{Conformal p-values.}
Let $n=|\Omega_{\mathrm{cal}}|$ and index calibration nodes as $\{1,\ldots,n\}$, and test nodes as $\{n+1,\ldots,n+m\}$ with $m=|\Omega_{\mathrm{test}}|$. For each test node $j\in\{1,\ldots,m\}$, define
\begin{equation}
p_j
=
\frac{
1+
\sum_{i=1}^{n}\mathbbm{1}\{J_i<\widehat{J}_{n+j}\}
+
U_j\sum_{i=1}^{n}
\mathbbm{1}\{J_i=\widehat{J}_{n+j}\}
}{
n+1
},
\label{eq:p_def}
\end{equation}
where $U_j\sim\mathrm{Unif}[0,1]$ is used for random tie-breaking. By the construction above, calibration nodes with $B_i=0$ cannot contribute to either sum. Hence~\eqref{eq:p_def} reduces exactly to the p-value construction in the main text:
\begin{equation}
p_j
=
\frac{
1+
\sum_{i:\,B_i=1}
\mathbbm{1}\{\hat{\kappa}(i)<\hat{\kappa}(n+j)\}
+
U_j
\sum_{i:\,B_i=1}
\mathbbm{1}\{\hat{\kappa}(i)=\hat{\kappa}(n+j)\}
}{
n+1
}.
\end{equation}

\paragraph{Generalized validity.}
Let
\begin{equation}
J_{n+j}=J(n+j,B_{n+j})
\end{equation}
denote the true score of test node $j$, and define its oracle randomized conformal p-value
\begin{equation}
p_j^\star
=
\frac{
\sum_{i=1}^{n}\mathbbm{1}\{J_i<J_{n+j}\}
+
U_j\left(
1+\sum_{i=1}^{n}\mathbbm{1}\{J_i=J_{n+j}\}
\right)
}{
n+1
}.
\label{eq:oracle_p}
\end{equation}
By exchangeability of calibration and test nodes (Assumption~\ref{assump:exchangeability}), the oracle conformal
rank $p_j^\star$ is super-uniform.

On the null event $H_j$, we have $B_{n+j}=1$, and hence
\[
J_{n+j}
=
J(n+j,1)
=
\widehat{J}_{n+j}.
\]
Comparing~\eqref{eq:p_def} and~\eqref{eq:oracle_p}, on $H_j$,
\begin{equation}
p_j
=
p_j^\star+\frac{1-U_j}{n+1}
\ge p_j^\star.
\end{equation}
Therefore, for every $t\in[0,1]$,
\begin{equation}
\mathbb{P}\big(p_j\le t,\ H_j\big)
\le
\mathbb{P}\big(p_j^\star\le t\big)
\le t.
\label{eq:generalized_validity}
\end{equation}
This is the generalized validity property for the random hypothesis $H_j$ in conformal selection \cite{jin2023selection}.

\paragraph{BH step and FDR control.}
Apply Benjamini--Hochberg (BH)
\cite{benjamini1995controlling} at level $\alpha$ to
$\{p_j\}_{j=1}^m$.
Let $p_{(1)}\le \cdots \le p_{(m)}$ and
\begin{equation}
k^\star
=
\max\Big\{
k:\ p_{(k)}\le \frac{\alpha k}{m}
\Big\},
\qquad
\mathcal{R}
=
\{j:\ p_j\le p_{(k^\star)}\},
\end{equation}
where $\mathcal{R}$ is the BH rejection set. In our setting, rejecting $H_j$ means declaring node $j$ reliable (i.e., deploying its imputation). Let $R=|\mathcal{R}|$ and define the number of false discoveries (deployed but truly risky) as
\begin{equation}
V
=
\sum_{j=1}^m
\mathbbm{1}\{j\in\mathcal{R},\ H_j\ \text{true}\}.
\end{equation}
The FDR is
\begin{equation}
\mathrm{FDR}
=
\mathbb{E}\!\left[
\frac{V}{\max\{R,1\}}
\right].
\end{equation}

For the multiple-testing guarantee, we use the exchangeable-data result for conformal selection in \cite{jin2023selection}. Specifically, assume that for each test node $j$, the score collection
\[
\{J_1,\ldots,J_n,J_{n+j}\}
\]
is exchangeable conditional on the remaining test threshold scores $\{\widehat{J}_{n+\ell}\}_{\ell\ne j}$, and that
\[
\{J_i\}_{i=1}^n
\cup
\{\widehat{J}_{n+\ell}\}_{\ell=1}^m
\]
has no ties almost surely.

Under the no-ties condition,~\eqref{eq:p_def} reduces to
\begin{equation}
p_j
=
\frac{
1+
\sum_{i=1}^{n}
\mathbbm{1}\{J_i<\widehat{J}_{n+j}\}
}{
n+1
},
\end{equation}
which is exactly the conformal p-value used in the exchangeable-data cfBH guarantee of \cite{jin2023selection}. Moreover, defining the reliability label as $1-B_i$ makes $J$ monotone in this label, and the risky-imputation null $H_j$ corresponds to evaluating the test score at the null label $B_j=1$ (equivalently, reliability label $1-B_j=0$).

Therefore, by the finite-sample FDR guarantee for cfBH under exchangeability \cite{jin2023selection},
\begin{equation}
\mathrm{FDR}
=
\mathbb{E}\!\left[
\frac{
\sum_{j=1}^{m}
\mathbbm{1}\{j\in\mathcal{R},\ H_j\ \text{true}\}
}{
\max\{|\mathcal{R}|,1\}
}
\right]
\le \alpha.
\end{equation}
Therefore, among the deployed imputations (BH rejections), the expected fraction of truly risky ones is controlled at level $\alpha$. This completes the proof.

%% file: sec/appendix_2.tex
\vspace{-5pt}
\section{Datasets Analysis}
\label{app:data}

\begin{table}[t]
\centering
\small
\caption{Statistics of the preprocessed datasets.}
\vspace{-10pt}
\label{tab:dataset-stats}
\scalebox{1}{ 
\begin{tabular}{lcc}
\toprule
\textbf{Dataset} & \textbf{\#Patient} & \textbf{\#Avg Event.}\\
\midrule
Mayo Clinic    & 84  & 7.40  \\
MIMIC-III   & 4243  & 2.20\\
MIMIC-IV  & 28880 & 4.95 \\

\bottomrule
\end{tabular}
}
\vspace{-10pt}
\end{table}

\paragraph{Dataset statistics.}
Table~\ref{tab:dataset-stats} reports the basic statistics of the three preprocessed datasets used in our study.
The Mayo Clinic dataset was collected from patients diagnosed with type~1 diabetes who had multiple clinical visits at Mayo Clinic during 2022--2023. We focus on routinely used metabolic and anthropometric measurements, including HbA1c (A1C), glucose, cholesterol and lipid panel components, and basic body measurements (weight, height, BMI), together with basic demographics.
MIMIC-III and MIMIC-IV are publicly available critical-care EHR datasets from the Beth Israel Deaconess Medical Center. MIMIC-III covers ICU stays between 2001 and 2012, while MIMIC-IV contains more recent ICU and emergency department records from 2008 to 2019.


\begin{figure}[t]
    \centering
    \includegraphics[width=0.85\linewidth]{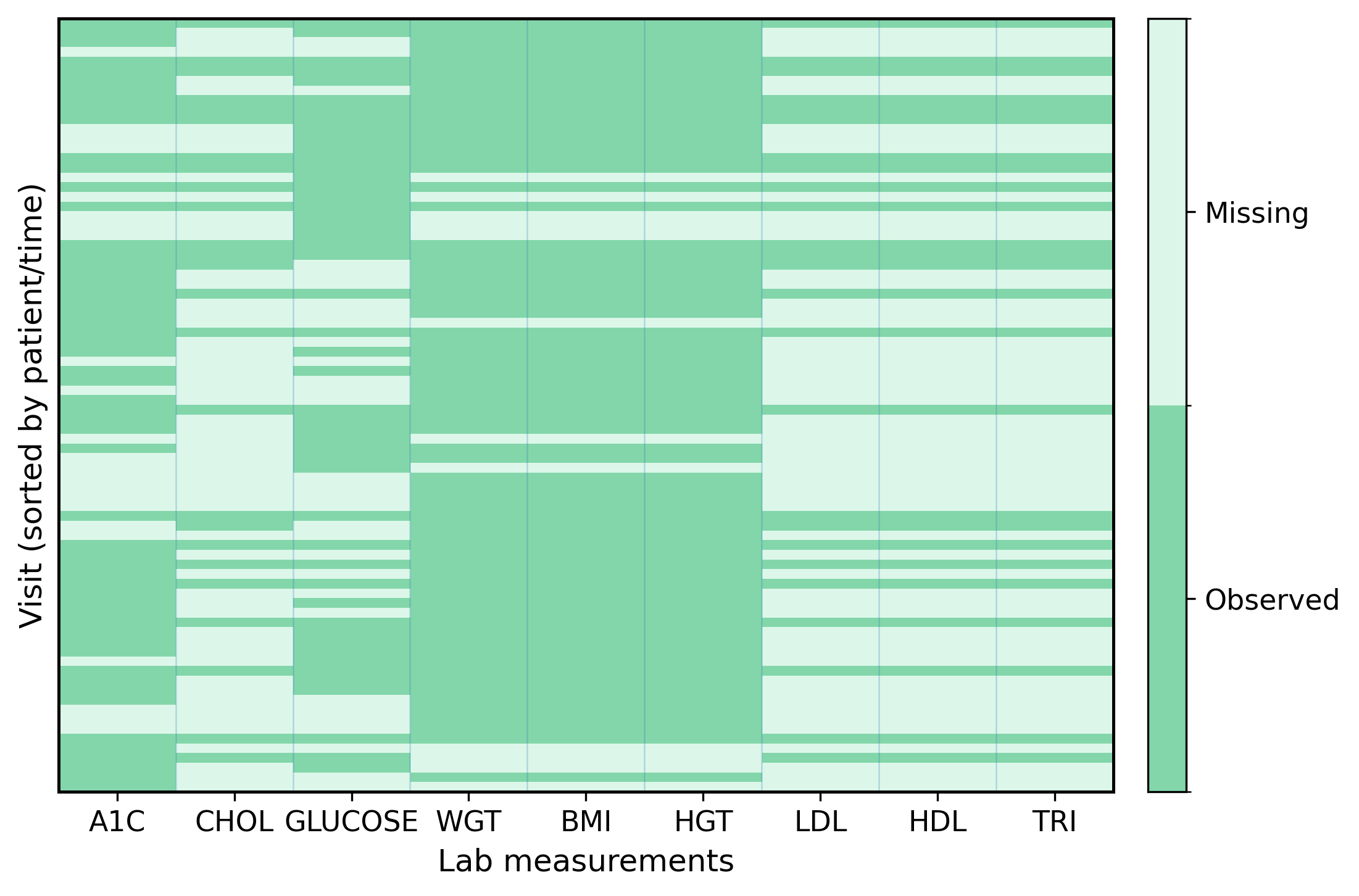}
    \vspace{-10pt}
    \caption{Rows are visits sorted by patient and time; columns are lab measurements. Darker cells indicate observed values, and lighter cells indicate missing values.}
    \vspace{-5pt}
    \label{fig:mayo-missing-heatmap}
\end{figure}


\begin{figure}[t]
    \centering
    \begin{subfigure}[t]{0.45\linewidth}
        \centering
        \includegraphics[width=\linewidth]{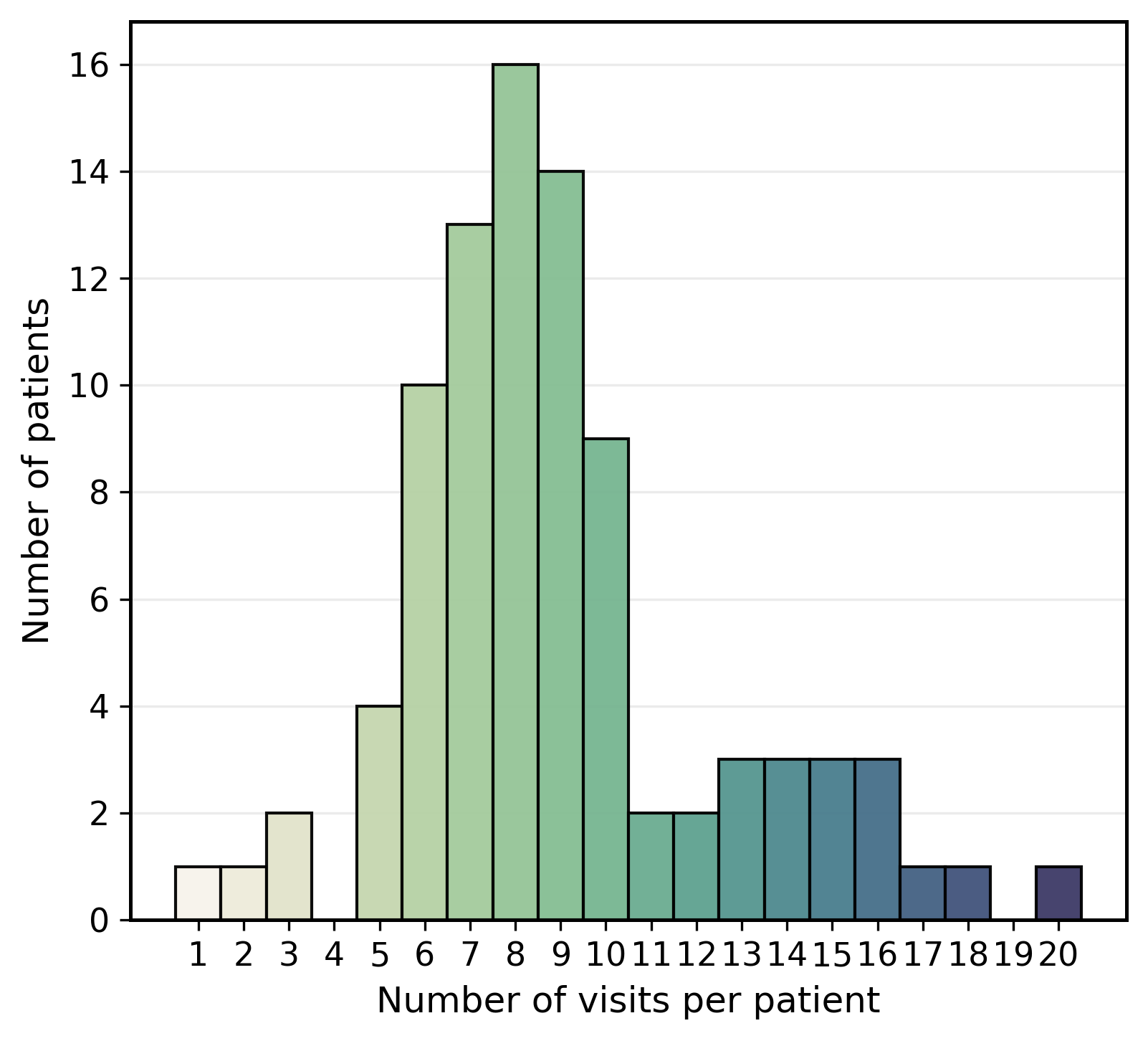}
        \label{fig:mayo-visit-dist}
    \end{subfigure}
    \hfill
    \begin{subfigure}[t]{0.45\linewidth}
        \centering
        \includegraphics[width=\linewidth]{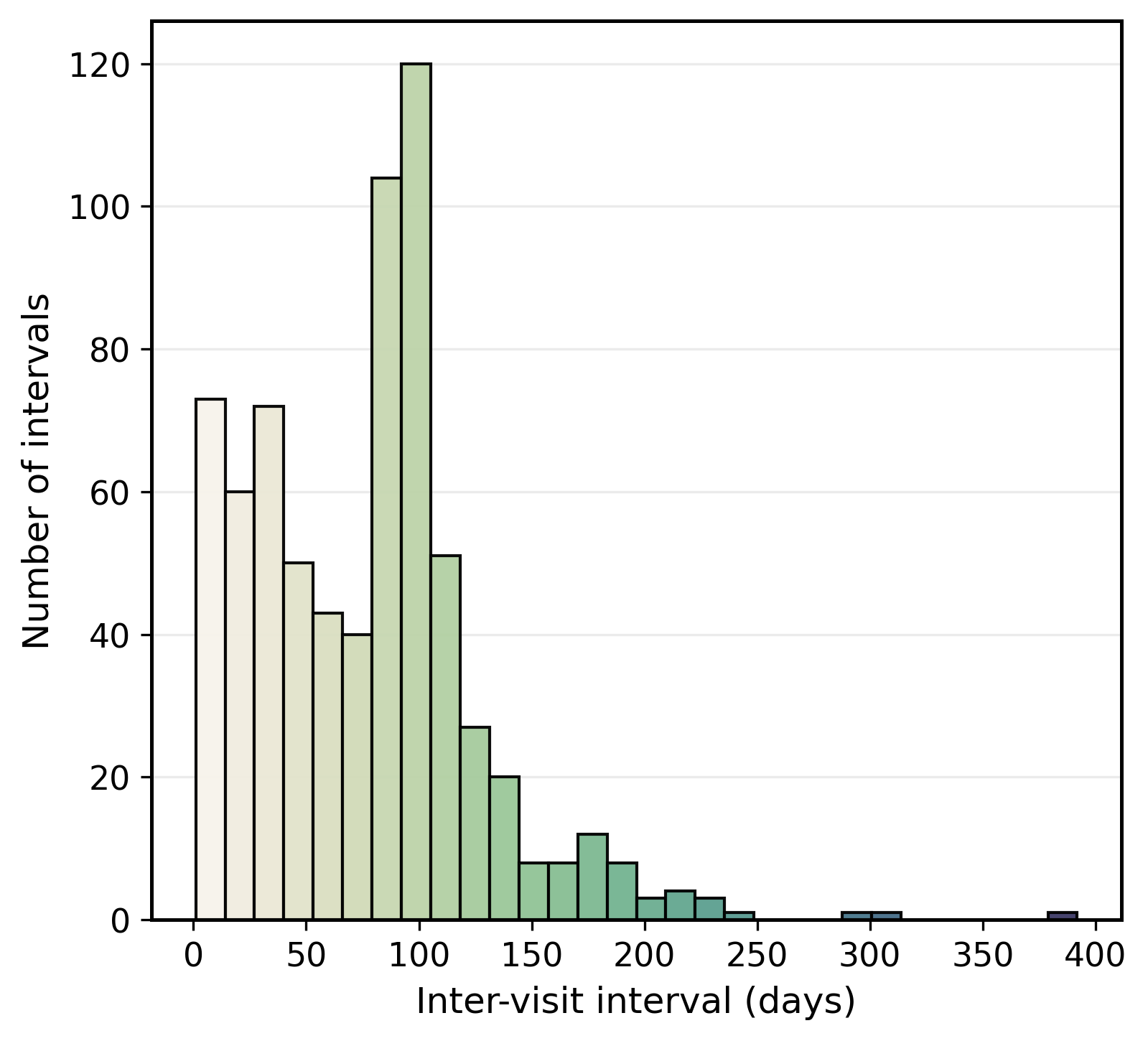}
        \label{fig:mayo-interval-dist}
    \end{subfigure}
    \vspace{-10pt}
    \caption{Left: distribution of the number of visits per patient. 
Right: distribution of inter-visit intervals between consecutive visits within each patient.}
\vspace{-10pt}
    \label{fig:mayo-dists}
\end{figure}

\begin{figure}[t]
    \centering
    \includegraphics[width=0.90\linewidth]{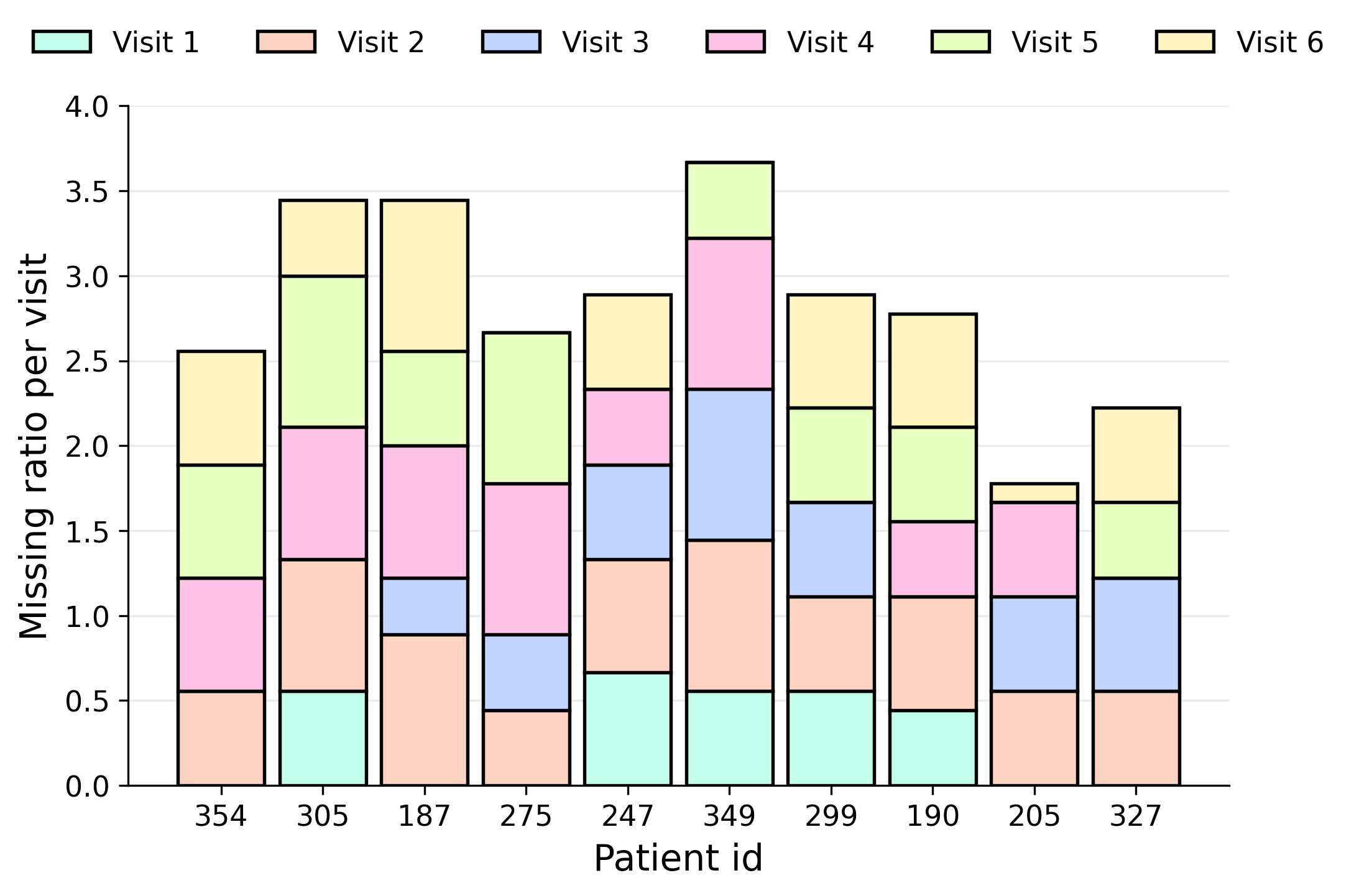}
    \vspace{-10pt}
    \caption{Visit-wise missingness for 10 randomly sampled Mayo Clinic patients. 
Stacked segments show missing ratios at each specific visit index, with total bar height indicating cumulative missingness.}
\vspace{-10pt}
    \label{fig:mayo-missratio-stacked}
    
\end{figure}

\paragraph{Missingness patterns across measurements.}
Figure~\ref{fig:mayo-missing-heatmap} visualizes missingness across visits and lab measurements in the Mayo Clinic data.
Rows correspond to visits sorted by patient and time, and columns correspond to measurements.
Missingness is highly heterogeneous: some measurements are recorded consistently, whereas lipid panel components such as LDL, HDL, and triglycerides show substantially higher missing rates.
This pattern reflects opportunistic test ordering in routine care.

\paragraph{Heterogeneous longitudinal follow-up.}
Figure~\ref{fig:mayo-dists} (left) shows the distribution of the number of visits per patient in Mayo Clinic data.
Most patients have a moderate number of visits, while a small fraction have substantially fewer or more visits.
This heterogeneity means that some patients provide limited intra-patient history, motivating the use of cross-patient information when individual trajectories are short.

\paragraph{Irregular inter-visit intervals.}
Figure~\ref{fig:mayo-dists} (right) reports inter-visit time gaps, in days, between consecutive visits within each patient in Mayo Clinic data.
The distribution is highly non-uniform and is consistent with periodic diabetes monitoring, while also showing substantial irregularity across patients.

\paragraph{Visit-wise missingness dynamics.}
Figure~\ref{fig:mayo-missratio-stacked} summarizes visit-wise missing ratios for 10 randomly sampled patients in Mayo Clinic data.
Missingness varies both across visits within the same patient and across patients at the same visit index, further highlighting the irregular and opportunistic nature of measurement collection in real-world clinical practice.

%% file: sec/appendix_5.tex
\vspace{-5pt}
\section{Evaluation Metrics}
\label{app:exp-details}
Let $\hat{y}_i$ denote the imputed value and $y_i$ denote the ground-truth value for an evaluated node $i$. 
The mean absolute error (MAE) is defined as
\begin{equation}
\mathrm{MAE}
=
\frac{1}{|\Omega_{\mathrm{test}}|}
\sum_{i \in \Omega_{\mathrm{test}}}
|\hat{y}_i - y_i| .
\end{equation}
The root mean squared error (RMSE) is
\begin{equation}
\mathrm{RMSE}
=
\sqrt{
\frac{1}{|\Omega_{\mathrm{test}}|}
\sum_{i \in \Omega_{\mathrm{test}}}
(\hat{y}_i - y_i)^2
}.
\end{equation}

For selective-release evaluation, let $\mathcal{S}$ denote the set of released imputations selected by the FDR-control procedure. 
Given the clinically meaningful error threshold $\delta$, an imputation is considered reliable if $|\hat{y}_i-y_i| < \delta$ and risky otherwise. 
We define precision as the fraction of released imputations that are truly reliable:
\begin{equation}
\mathrm{Precision}
=
\frac{
|\{i \in \mathcal{S}: |\hat{y}_i-y_i| < \delta\}|
}{
\max\{|\mathcal{S}|,1\}
}.
\end{equation}
Equivalently, since empirical FDR measures the fraction of risky imputations among the released set, precision corresponds to $1-\mathrm{FDR}$. Higher precision indicates that a larger fraction of released imputations are within the clinically acceptable error range.

%% file: sec/appendix_4.tex
\begin{figure}[t]
    \centering
    \begin{subfigure}{0.48\linewidth}
        \centering
        \includegraphics[width=\linewidth]{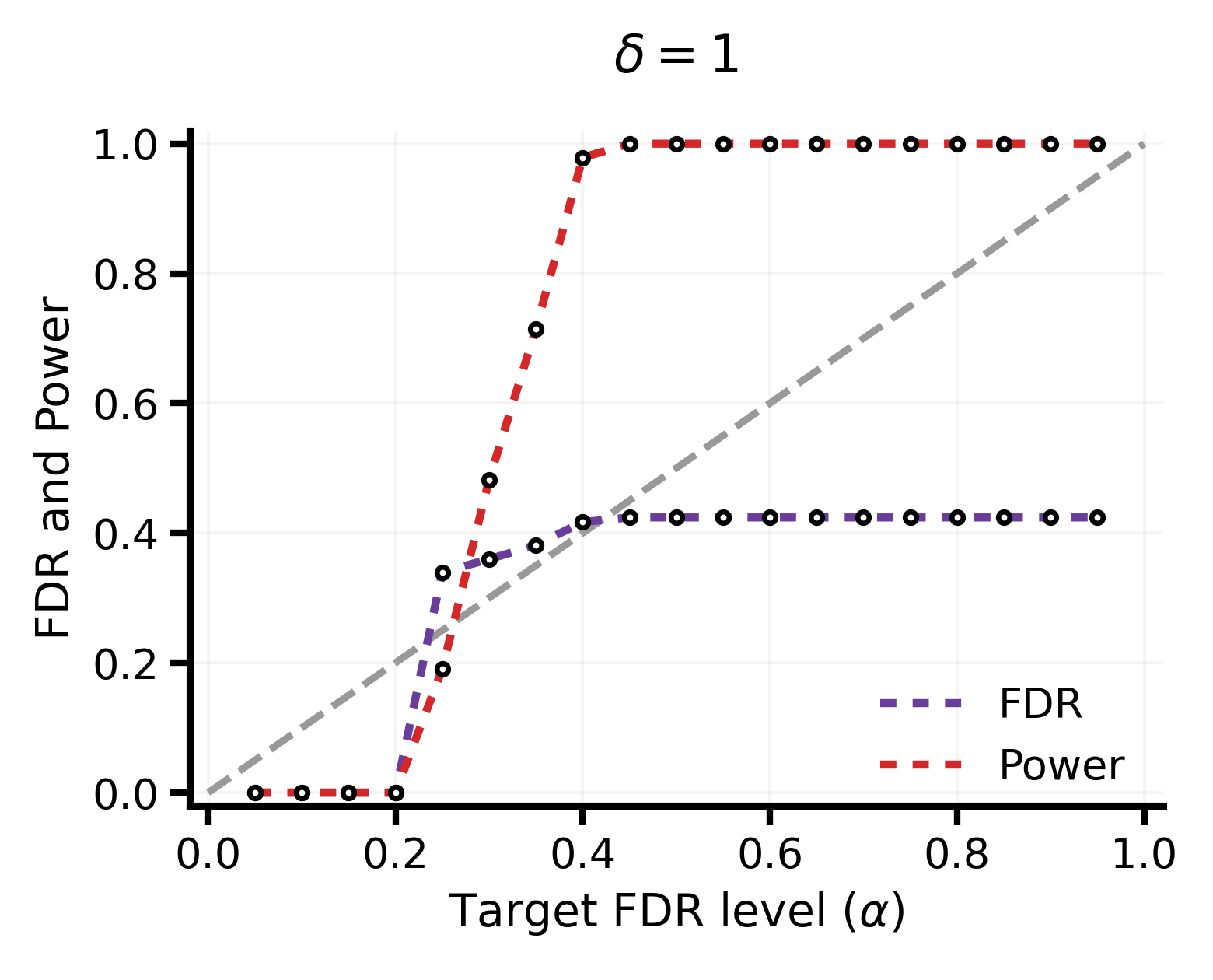}
        \caption{MIMIC-III}
    \end{subfigure}
    \hfill
    \begin{subfigure}{0.48\linewidth}
        \centering
        \includegraphics[width=\linewidth]{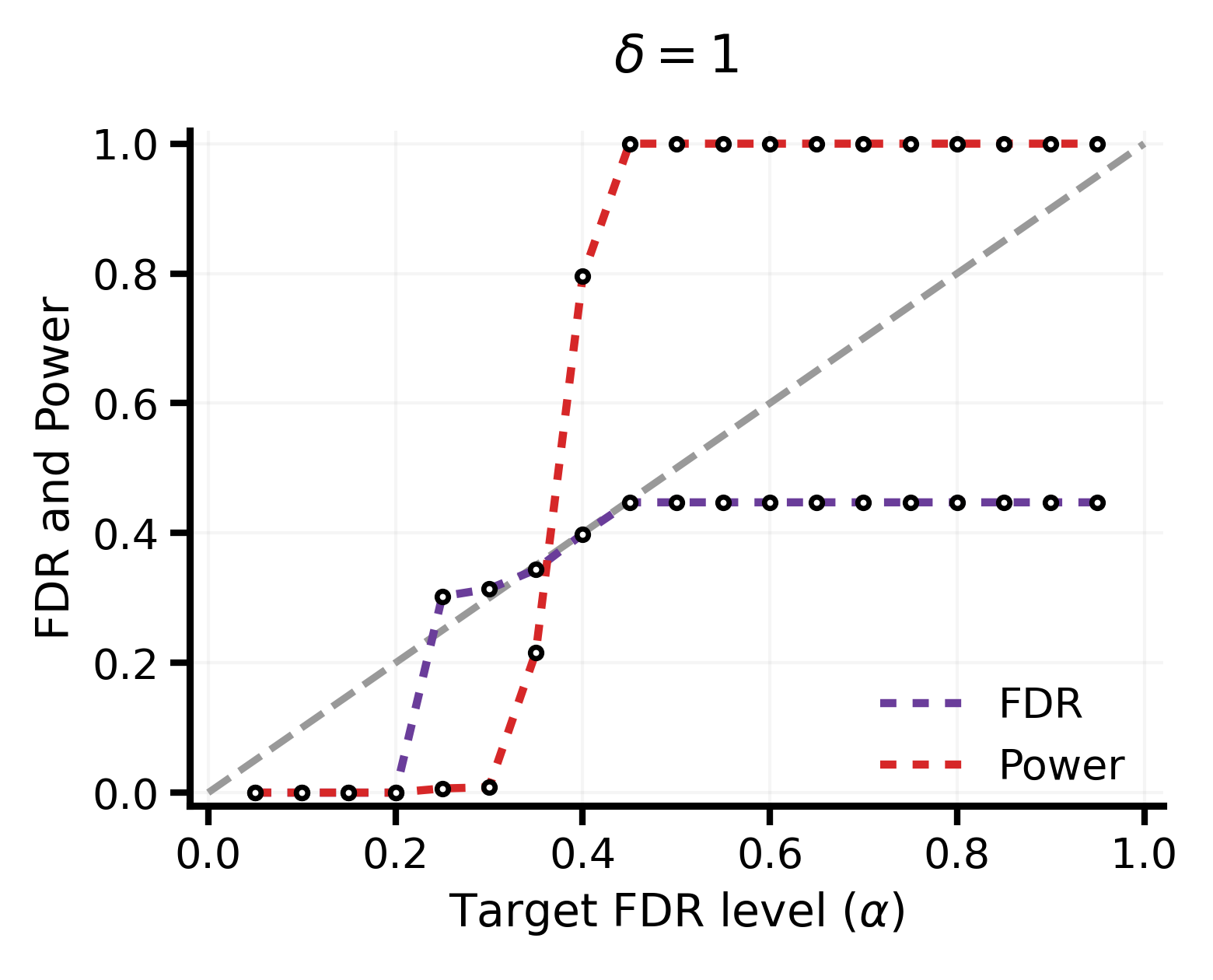}
        \caption{MIMIC-IV}
    \end{subfigure}
    \caption{Additional FDR--power trade-off curves on MIMIC-III and MIMIC-IV with $\delta=1.0$.}
    \label{fig:fdr_mimic_appendix}
\end{figure}

\section{FDR Curves on MIMIC-III and MIMIC-IV}
\label{app:fdr_power_mimic}

Figure~\ref{fig:fdr_mimic_appendix} provides additional FDR--power trade-off curves on MIMIC-III and MIMIC-IV. 
We fix the clinically interpretable tolerance to $\delta=1.0$ and sweep the target FDR level $\alpha$. 
These results show how the BH target level controls the selective release behavior on public clinical datasets: increasing $\alpha$ makes the selection rule less restrictive and generally increases power, while the achieved FDR remains interpretable relative to the target level.